\documentclass[runningheads]{llncs}
\usepackage[T1]{fontenc}
\usepackage[table]{xcolor}
\definecolor{CColor}{rgb}{0.01,0.31,0.59}
\definecolor{GGray}{rgb}{0.80,0.90,1}
\definecolor{Shady}{rgb}{0.9,0.9,0.9}
\definecolor{kaistblue}{RGB}{20,135,200}
\definecolor{kaistdarkblue}{RGB}{0,65,145}
\definecolor{urbanablue}{RGB}{19,41,75}
\definecolor{urbanaorange}{RGB}{232,74,39}
\definecolor{drp}{rgb}{0.53,0.15,0.34}
\definecolor{maroon}{cmyk}{0,0.87,0.68,0.32}

\usepackage{hyperref}       
\usepackage{url}            
\usepackage{booktabs}       
\usepackage{amsfonts}       
\usepackage{nicefrac}       
\usepackage{microtype}      
\usepackage{xcolor}         
\usepackage{comment}
\usepackage{amsthm}
\usepackage[shortlabels]{enumitem}
\usepackage{float}
\usepackage{bbm}
\usepackage{mathtools}
\usepackage{multirow}
\usepackage{diagbox}
\usepackage{comment}

\usepackage{rotating}
\usepackage[normalem]{ulem}
\usepackage{bm}
\usepackage{subfig}
\usepackage{booktabs, makecell, tabularx}
\usepackage{rotating}
\usepackage[export]{adjustbox}
\usepackage[misc]{ifsym}

\newcommand{\tabincell}[2]{\begin{tabular}{@{}#1@{}}#2\end{tabular}}
\theoremstyle{remark}
\usepackage{algorithm}
\usepackage{algorithmic}

\usepackage{wrapfig}
\usepackage{tikz}
\usepackage{pifont}
\newcommand*\circled[1]{\tikz[baseline=(char.base)]{\node[shape=circle,draw,inner sep=0.4pt] (char) {#1};}}
\newcommand{\gr}{\rowcolor[gray]{.95}}

\begin{document}
\title{Enhancing Adversarial Training via Reweighting Optimization Trajectory}
\toctitle{Enhancing Adversarial Training via Reweighting Optimization Trajectory} 
%
%

\author{Tianjin Huang  \Letter \inst{1} \thanks{These authors contributed equally to this research. Accepted by ECML 2023.}  \and
Shiwei Liu \inst{2}$^*$  \and Tianlong Chen\inst{2}$^*$ \and Meng Fang\inst{3,1} \and Li Shen\inst{4} \and Vlado Menkovski\inst{1} \and Lu Yin\inst{1}\and Yulong Pei\inst{1} \and Mykola Pechenizkiy\inst{1}}

\authorrunning{T. Huang et al.}

\institute{Eindhoven University of Technology, Eindhoven, the Netherlands \\ 
\email{\{t.huang,y.pei.1,v.menkovski,m.penchenizkiy,l.yin\}@tue.nl}  
\and University of Texas at Austin, Austin, USA  \\
\email{tianlong.chen@utexas.edu, shiwei.liu@austin.utexas.edu}
\and University of Liverpool, Liverpool, UK  \\
\email{mfang@liverpool.ac.uk}
\and JD Explore Academy, Beijing, China  \\
\email{mathshenli@gmail.com}
}
\tocauthor{Tianjin Huang, Shiwei Liu, Tianlong Chen, Meng Fang, Li Shen, Vlado Menkovski, Lu Yin, Yulong Pei, and Mykola Pechenizkiy}  
\maketitle              
%

\begin{abstract}
Despite the fact that adversarial training has become the de facto method for improving the robustness of deep neural networks, it is well-known that vanilla adversarial training suffers from daunting robust overfitting, resulting in unsatisfactory robust generalization. A number of approaches have been proposed to address these drawbacks such as extra regularization, adversarial weights perturbation, and training with more data  over the last few years. However, the robust generalization improvement is yet far from satisfactory. In this paper, we approach this challenge with a brand new perspective -- refining historical optimization trajectories. We propose a new method named \textbf{Weighted Optimization Trajectories (WOT)} that leverages the optimization trajectories of adversarial training in time. We have conducted extensive experiments to demonstrate the effectiveness of WOT under various state-of-the-art adversarial attacks. Our results show that WOT integrates seamlessly with the existing adversarial training methods and consistently overcomes the robust overfitting issue, resulting in better adversarial robustness. For example, WOT boosts the robust accuracy of AT-PGD under AA-$L_{\infty}$ attack by 1.53\% $\sim$ 6.11\% and meanwhile increases the clean accuracy by 0.55\%$\sim$5.47\%  across SVHN, CIFAR-10, CIFAR-100, and Tiny-ImageNet datasets. The code is available at \url{https://github.com/TienjinHuang/WOT.}
\end{abstract}
\section{Introduction}\label{intro}
Deep neural networks (DNNs) have achieved enormous breakthroughs in various fields, e.g., image classification~\cite{Hinton2012,He2016}, speech recognition~\cite{Hinton2012}, object detection~\cite{Girshick2014} and etc. However, it has been shown that they are vulnerable to adversarial examples, i.e., carefully crafted imperceptible perturbations on inputs can easily change the prediction of the model~\cite{Szegedy2013,Goodfellow2014}. The vulnerability of DNNs hinders their applications in risk-sensitive tasks such as face recognition, autonomous driving, and medical diagnostics. While various methods have been proposed to obtain robustness against adversarial perturbations, adversarial training~\cite{madry2017towards} is the leading approach  to achieve adversarial robustness. 

However, the vanilla adversarial training usually suffers from daunting robust overfitting, resulting in poor robust generalization\footnote{Robust generalization refers to the gap between the adversarial accuracy of the training set and test set, following previous work~\cite{chen2020robust,wu2020adversarial,stutz2021relating}.}~\cite{rice2020overfitting}. To tackle this issue, a number of methods from different perspectives have been proposed including but not limited to training with more data~\cite{schmidt2018adversarially,rebuffi2021fixing,sehwag2021robust,carmon2019unlabeled,alayrac2019labels}, adversarial weights perturbation~\cite{wu2020adversarial,yu2021robust}, and knowledge distillation and stochastic weights averaging (SWA)~\cite{chen2020robust}. Recently, ~\cite{stutz2021relating} empirically show that the improved adversarial robustness can be attributed to the flatter loss landscape at the minima.

\begin{figure*}[tbh]
\vskip -0.1 in
\centering
\includegraphics[width=0.32\textwidth]{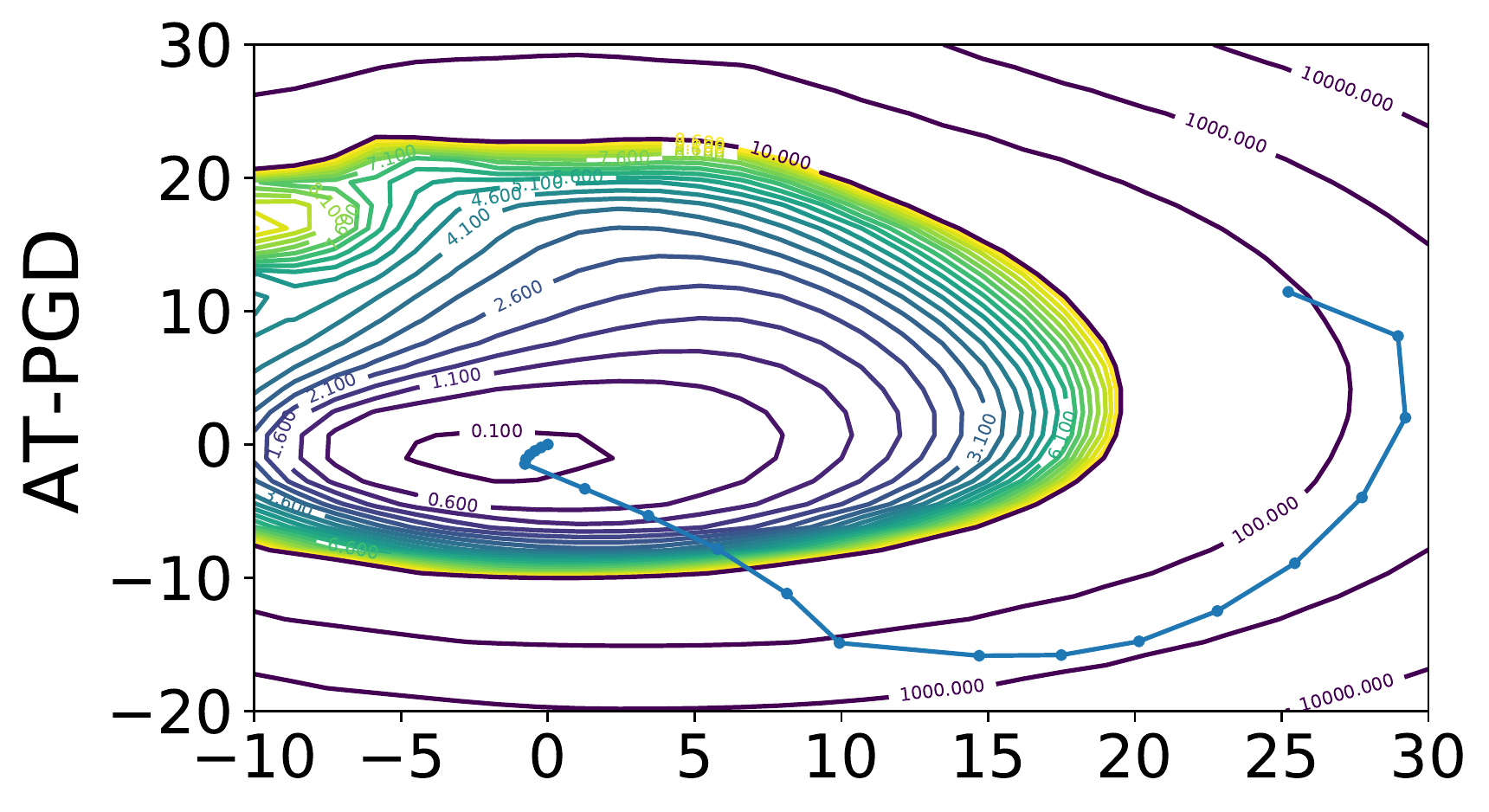}
\includegraphics[width=0.32\textwidth]{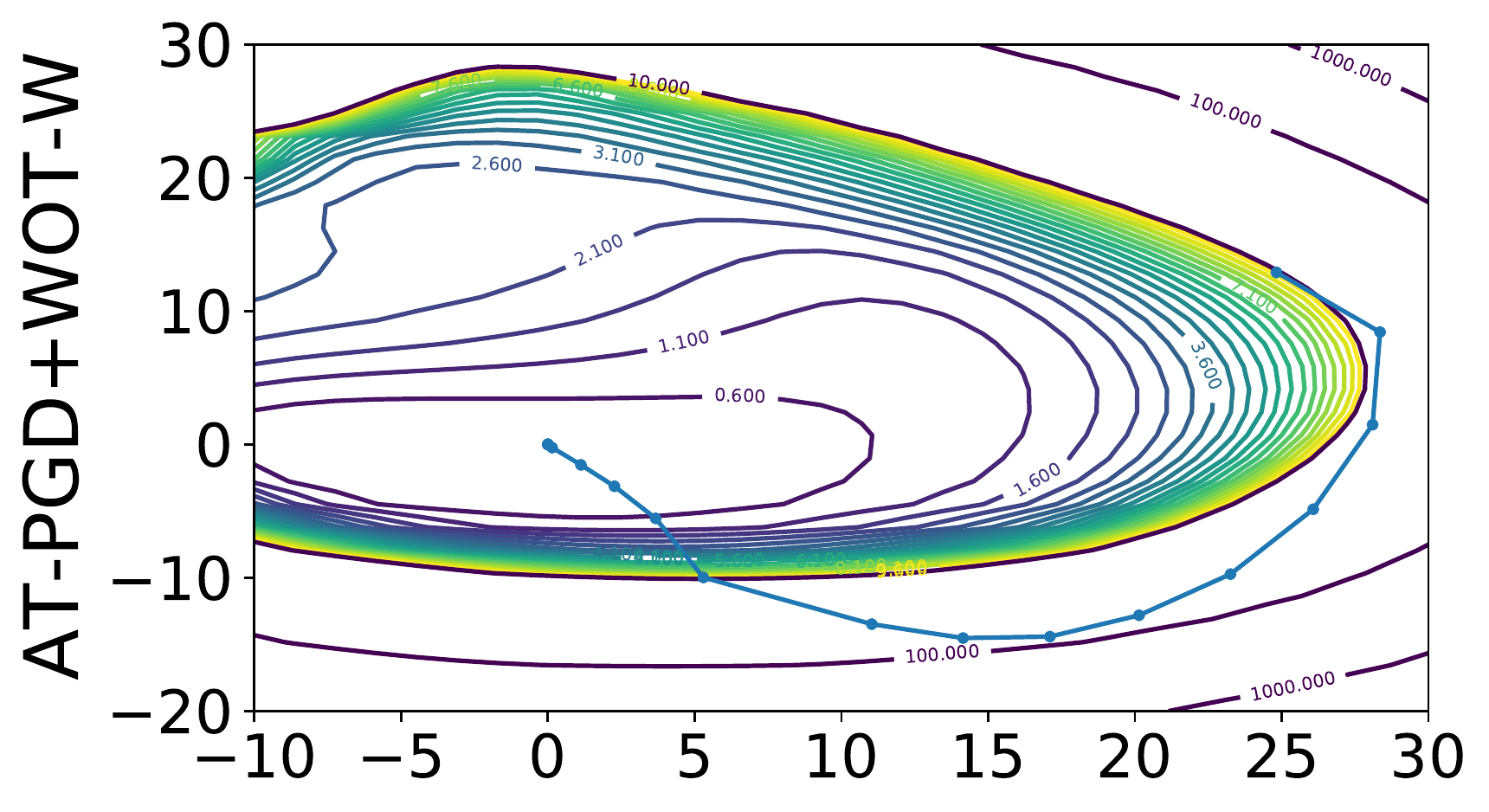}
\includegraphics[width=0.32\textwidth]{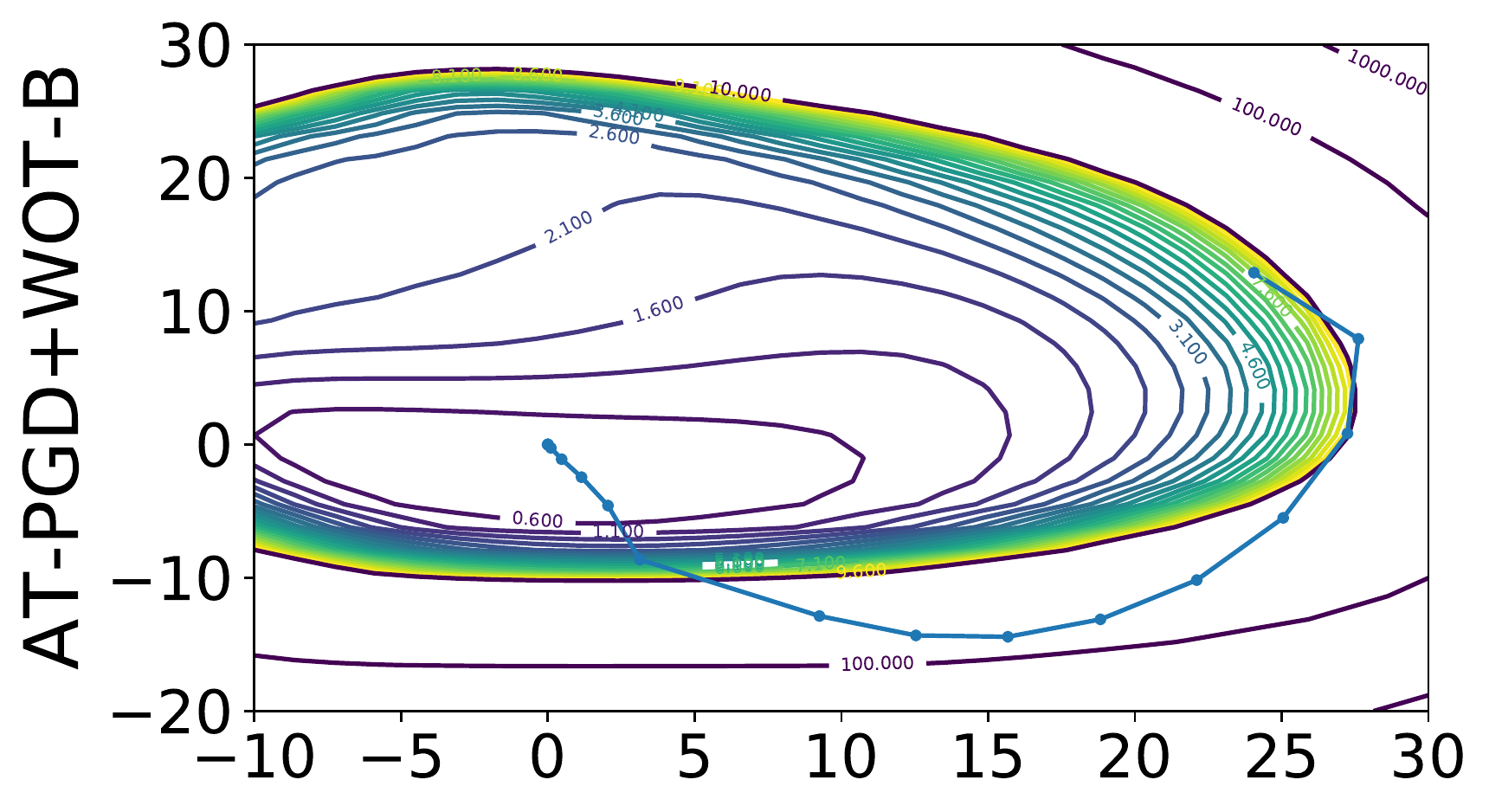}
\caption{Visualization of loss contours and optimization trajectories for AT-PGD, AT-PGD+WOT-W, and AT-PGD+WOT-B, respectively. The experiments are conducted on CIFAR-10 with PreRN-18.}
\label{fig:contour_traj}
\end{figure*}

Although the generalization properties of SGD-based optimizer under standard training setting have been well studied~\cite{zhang2017musings,elisseeff2005stability,zhou2018generalization,hardt2016train}, the corresponding robust generalization property under adversarial setting has not been fully explored. Among previous studies,~\cite{chen2020robust} heuristically adopts stochastic weight averaging (SWA) and average model weights along the optimization trajectory, which  potentially mitigates robust overfitting. However, it has been shown that naive weight averaging is not general enough to fundamentally address this problem, still prone to
robust overfitting~\cite{rebuffi2021fixing}.   
Instead of simply averaging weights, we propose a new approach - \textbf{Weighted Optimization Trajectories} (briefly \textbf{WOT)} for the first time showing that we can largely improve the flatness of solutions of existing adversarial training variants by periodically refining a set of historical optimization trajectories. Compared with the existing approaches, our method has three unique design contributions: \ding{182} our refinement is obtained by maximizing the robust accuracy on the \underline{unseen hold-out set}, which is naturally advantageous to address the overfitting issue; \ding{183} our refinement is performed on a set of previous optimization trajectories rather than solely on previous weights; \ding{184} we further propose a block-wise trajectory refinement, which significantly enlarges the optimization space of refinement, leading to better robust performance. We conduct rigorous experiments to demonstrate the effectiveness of these design novelties in Section~\ref{superioriRA} as well as the ablation study in Section~\ref{sec:ablation}. Simple as it looks in Figure~\ref{fig:contour_traj}, the optimization trajectories after refining converge to a flatter loss valley compared to the vanilla AT-PGD, indicating the improved robust generalization~\cite{wu2020adversarial,wu2020revisiting,stutz2021relating}. 

Extensive experiments on different architectures and datasets show that WOT seamlessly mingles with the existing adversarial training methods with consistent robust accuracy improvement. For example, WOT-B directly boosts the robust accuracy over AT-PGD (early stops) under AA-$L_{\infty}$ attack by 6.11\%, 1.53\%, 1.57\%, and 4.38\% on SVHN, CIFAR-10, CIFAR-100, and Tiny ImageNet, respectively; meanwhile improves the corresponding clean accuracy by 0.55$\% \sim$ 5.47\%. Moreover, we show that WOT can completely prevent robust overfitting across different attack approaches, including the strongest one off-the-shelf - AA-$L_{\infty}$ attack.

\section{Related Work}
\textbf{Adversarial Attacks.} Adversarial examples were first illustrated in~\cite{Szegedy2013}. Following~\cite{Szegedy2013}, many adversarial attacks have been proposed and can be categorized into white-box and black-box attacks. White-box attacks have full access to the model when crafting adversarial examples. Popular white-box attacks are FGSM attack~\cite{Goodfellow2014}, PGD attack~\cite{Madry2017}, Deepfool~\cite{Moosavi-Dezfooli2016} and C\&W~\cite{carlini2017towards}. Black-box attacks generate adversarial examples without any knowledge of the model. They are query-based attacks, e.g., SPSA attack~\cite{uesato2018adversarial}, Square attack~\cite{andriushchenko2020square}, and transferability-based attacks, e.g., DIM~\cite{xie2019improving}, TIM~\cite{dong2019evading} and DA attack~\cite{huang2022direction}. Recently, ~\cite{croce2020reliable} proposed Autoattack (AA) for reliable adversarial robustness evaluation which is an ensembled adversarial attack containing white-box and black-box attacks.  AA attack has been recognized as the most reliable method for evaluating the model's adversarial robustness~\cite{croce2020reliable} and will be used as the main evaluation method in this paper. \\  
\textbf{Adversarial Robustness.} Many methods have been proposed to improve the model's robustness such as gradient regularization~\cite{ross2018improving}, curvature regularization~\cite{Moosavi-Dezfooli2018,huang2020bridging}, randomized smoothing~\cite{cohen2019certified}, local linearization~\cite{Qin2019}, adversarial training methods~\cite{Goodfellow2014,Madry2017,zhang2019theoretically,wu2020adversarial,wang2019improving,zhang2020attacks,huang2021calibrated,balaji2019instance,zhang2020geometry,pang2022robustness}, and etc. Among all these methods, adversarial training has been the de facto method for achieving adversarial robustness. We briefly introduce four commonly used adversarial training methods that we use as baselines in this study. 

Given a $C$-class dataset $S=\{(x_{i},y_{i})|x_{i}\in \mathbb{R}^{d}, y_{i} \in \mathbb{R}\}^{n}_{i=1}$, the cross-entropy loss $L(\cdot)$ and the DNN function $f_{w}:\mathbb{R}^{d}\xrightarrow{}\mathbb{R}^{C}$.\\ \textbf{AT-PGD~\cite{Madry2017}} is formalized as \emph{min-max} optimization problem.
\begin{small}
\begin{align}
  \min_{w} \rho^{AT}(w),\;\;\rho^{AT}(w)=  \mathbb{E}_{(x,y)\sim S} \{\max_{\lVert \Delta x \rVert \leq \epsilon } L(f_{w}(x_i+\Delta x),y_i)\}, \notag
\end{align}
\end{small}
  where the \emph{inner maximization} finds the adversarial examples and $\epsilon$ is the allowed perturbation magnitude. We  use by default AT to denote AT-PGD in the following sections.\\
\textbf{Trades~\cite{zhang2019theoretically}} separates training loss into  a cross-entropy loss ($CE$) and Kullback-Leibler ($\mathbf{KL}$) divergence loss to control clean accuracy and adversarial robustness respectively.
\begin{small}
\begin{align}
\rho^{TRADES}(w)=  \mathbb{E}_{(x,y)\sim S} \{CE(f_{w}(x_{i},y_i)+\beta \cdot \max_{\|\Delta x\|\leq \epsilon}\mathbf{KL}(f_{w}(x_{i})||f_{w}(x_{i}+\Delta x)\} \notag
\end{align}
\end{small}
\\\textbf{MART~\cite{wang2019improving}} designs the training loss as the binary cross-entropy loss ($BCE$) and an explicit regularization for misclassified examples. 
\begin{small}
\begin{align}
\small
\rho^{MART}(w)=  \mathbb{E}_{(x,y)\sim S} \{BCE(f_{w}(x_{i}+\Delta x,y_i) \\ \notag+\lambda \cdot \mathbf{KL}(f_{w}(x_{i})||f_{w}(x_{i}+\Delta x))\cdot(1-[f_{w}(x_{i})]_{y_{i}})\} \notag
\end{align}
\end{small}
\\\textbf{Adversarial Weights Perturbation (AWP)~\cite{wu2020adversarial}}  explicitly flattens the loss landscape by injecting the worst weight perturbations. 
\begin{small}
\begin{align}
\rho^{AWP}(w)= \max_{v \in \mathcal{V}} \mathbb{E}_{(x,y)\sim S} \max_{\|\Delta x\|\leq \epsilon} CE(f_{w+v}(x_{i}+\Delta x),y_i) \notag
\end{align}
\end{small}
\\ \textbf{Robust Overfitting and its Mitigation.}  \cite{rice2020overfitting} first identified the robust overfitting issue in AT that robust accuracy in test set degrades severely after the first learning rate decay and found that early stop is an effective strategy for mitigating the robust overfitting issue. Following~\cite{rice2020overfitting}, several studies have been proposed to explain and mitigate the robust overfitting issue~\cite{wu2020adversarial,singla2021low,chen2020robust,dong2021exploring,chen2022sparsity,stutz2021relating}. ~\cite{chen2020robust} showed that stochastic weight averaging (SWA) and knowledge distillation can mitigate the robust overfitting issue decently and ~\cite{singla2021low} found that low curvature activation function helps to mitigate the robust overfitting problem. ~\cite{dong2021exploring} took a step further to explain that the robust overfitting issue may be caused by the memorization of hard samples in the final phase of training. ~\cite{wu2020adversarial,yu2021robust,stutz2021relating} demonstrated that a flat loss landscape improves robust generalization and reduces the robust overfitting problem, which is in line with the sharpness studies in standard training setting~\cite{foret2020sharpness,jiang2019fantastic,dziugaite2017computing}. 

\section{Methodology}
\begin{wrapfigure}{r}{0.5 \linewidth}
 \vspace{1.em}
\begin{center}
\includegraphics[width=0.9\linewidth]{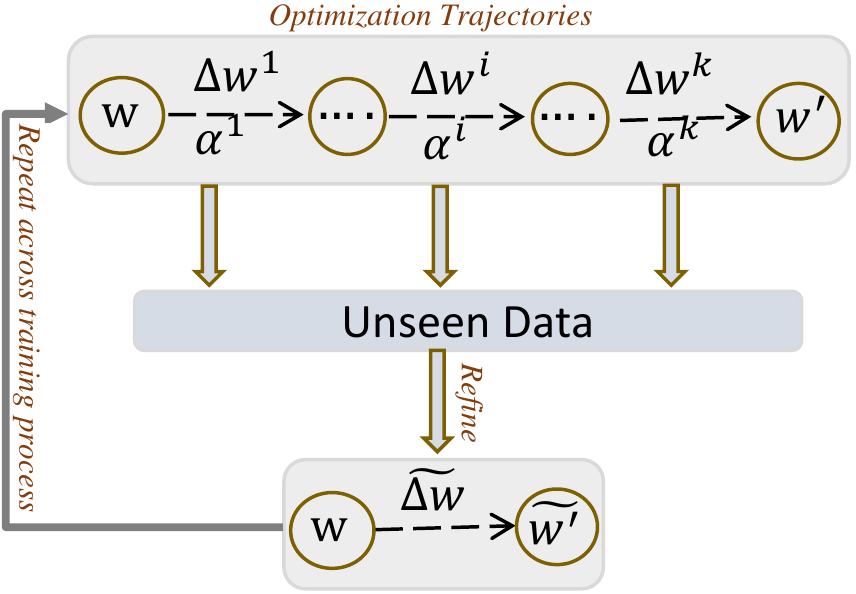}
\end{center}
\vspace{-2mm}
\caption{Sketch map of WOT.}
\label{fig:sketch_map}
 \vspace{-0mm}
\end{wrapfigure}

In this section, we will introduce \textbf{weighted optimization trajectories (WOT)}, a carefully designed method that refines the optimization trajectory of adversarial training towards a flatter region in the training loss landscape, to avoid robust overfitting.
Specifically, WOT collects a set of historical optimization trajectories and further learns a weighted combination of them  explicitly on the unseen set. The sketch map of WOT is shown in Figure~\ref{fig:sketch_map}. Concretely, WOT contains two steps: (1) collect optimization trajectories of adversarial training. (2) re-weight collected optimization trajectories and optimize weights according to the robust loss on an unseen set. Two unanswered problems of this process are how to collect optimization trajectories and how to construct the objective function of optimizing weights. We give detailed solutions as follows.   

\subsection{WOT: Optimization Trajectories}
\vspace{-1mm}
We denote optimization trajectories as the consecutively series status of weights in weight space after $n$ steps optimization.
Formally, given a deep neural network $f$ with the parameter $w \in \mathcal{W}$.  $n$ steps of optimization trajectories of adversarial training are denoted as $\{w^{1},w^{2}...w^{i}$,  $...,w^{n}\}$ where $w^{i}$ is the weight after the $i$-th optimization step. This process can also be simplified as follows: $\{w^{1},\Delta w^{1}...\Delta w^{i}\;,...,\;\Delta w^{n-1}\}$ 
where $\Delta w^{i}=w^{i+1}-w^{i}$.  In practice, it is time-consuming and space-consuming to collect the weights of each batch optimization step and it is also not necessary to collect the weights at a high frequency (see details in Figure~\ref{fig:impactofhyper}).  Therefore, we propose to collect weights for every $m$ batch optimization step  and the collected trajectories with $n$ optimization steps are re-denoted as follows:
\begin{small}
\begin{align}
    \Delta W=\{w^{1},\Delta w^{1}, ..., \Delta w^{i},...,\Delta w^{k}\},
\end{align}
\end{small}
where $k=\frac{n}{m}$. For brevity, we call $m$ the Gaps that control the length between two consecutive collections and $k$ the number of Gaps that controls the number of weights that are collected. 
\subsection{WOT: Objective Function}
We design the objective function based on historical optimization trajectories of model training. From the description of optimization trajectories introduced above, the weights $w^{\prime}$ with $n$ batch optimization steps from $w$ can be written as $w^{\prime}=w+\Delta w^{1}+...+\Delta w^{i}+..+\Delta w^{k}$. Since WOT refines the optimization trajectories by re-weighting them, the new model weights $\widetilde{w^{\prime}}$ after refining optimization trajectories can be expressed as follows:
\begin{small}
\begin{align}
\widetilde{w^{\prime}}=w+\widetilde{\Delta w}, \;\;
\widetilde{\Delta w}= \alpha^{1}\Delta w^{1}+...+\alpha^{i}\Delta w^{i}+...+\alpha^{k}\Delta w^{k}, \label{newdirection}
\end{align}
\end{small}
where $\alpha^{1},...,\alpha^{i},...,\alpha^{k}$ are optimizable variables. Considering that we expect to find the model with better robust generalization via optimizing $\alpha$, a straightforward idea is to optimize $\alpha^{i}$ with respect to improving its robust performance on a small unseen set. The philosophy for this idea is that if a model can generalize robustness better to an unseen set, it would probably generalize robustness well to the target unseen set.

Formally, the \textbf{objective function} of optimizing $\alpha^{i}$ is defined as follows:
\begin{align}
      \min_{0 \leq \alpha^{i}\leq 1} \max_{\|\Delta x_{uns}\|\leq \epsilon}{L(f_{w+\widetilde{\Delta w}}(x_{uns}+\Delta x_{uns}),y_{uns})},
      \label{wot-obj}
\end{align}
where ($x_{uns},y_{uns}$) is from an unseen set and $\Delta x_{uns}$ is the corresponding adversarial perturbations. We constrain $\alpha^{i}$ to [0,1] (see Appendix~F~\footnote{Appendix can be found in ~\url{https://arxiv.org/pdf/2306.14275v2.pdf}} for the results of other constraints for $\alpha^{i}$).
\textbf{Update $\alpha^{i}$}. $\alpha^{i}$ can be optimized by any SGD-based optimizers according to the objective function (Eq.~\ref{wot-obj}) described above. In this study, we update $\alpha^{i}$ by SGD optimizer with momentum buffer.
\begin{align}
    &m^{t}=m^{t-1} \cdot   \gamma+\nabla_{\alpha^{i}}{L(f_{w^{i-1}+\widetilde{\Delta w}}(x_{uns}+\Delta x_{uns}),y_{uns})} \label{update_alpha_m} \\
    &\alpha^{i}=\alpha^{i}-lr\cdot m^{t},\label{update_alpha}
\end{align}
where $m^{t}$ is the momentum buffer of $\alpha^{i}$ at the $t$-th step and $lr$ is the learning rate.

\begin{table*}[!h]
\caption{Robust accuracy of WOT under multiple adversarial attacks with various adversarial training variants. The experiments are conducted on CIFAR-10 with the PreRN-18 architecture. The best results are marked in bold. }
\begin{center}
\begin{sc}
\begin{tabular}{l|cccccc}
\toprule
Models  &FGSM&PGD-20 & PGD-100&CW$_{\infty}$&AA-$L_{\infty}$\\
\midrule
AT+early stop&57.30 &52.90 &51.90&50.90&47.43\\
AT+SWA &58.89&53.02&51.86&52.32&48.61\\
\gr AT+WOT-W ({Ours}) &58.50 &53.19& 51.90 &51.74 &48.36\\
\gr AT+WOT-B ({Ours})  &\textbf{59.67}&\textbf{54.85} &\textbf{53.77} &\textbf{52.56}&\textbf{48.96}\\
\midrule
Trades & 58.16&53.14 &52.17 &51.24&48.90 \\
Trades+SWA &58.07&53.17&52.22&50.91&49.07\\
\gr Trades+WOT-W ({Ours})  &\textbf{58.95} &\textbf{54.07}&\textbf{53.29} &51.74&49.95\\
\gr Trades+WOT-B ({Ours})  &58.50 &53.73 &52.95&\textbf{52.12}&\textbf{50.19}\\
\midrule
MART &  59.93&54.07&52.30&50.16 &47.01\\
MART+SWA & 58.19&54.21&53.56&49.39&46.86\\
\gr MART+WOT-W ({Ours}) &58.13 &53.79 &52.66&50.24 &47.43\\
\gr MART+WOT-B ({Ours}) &\textbf{59.95} &\textbf{55.13} &\textbf{54.09} &\textbf{50.56}&\textbf{47.49}\\
\midrule
AT+AWP   &59.11&55.45&54.88 &52.50&49.65\\
AT+AWP+SWA  &58.23&55.54&54.91&51.88&49.39\\
\gr AT+AWP+WOT-W ({Ours}) &59.05 &\textbf{55.95}&54.96 &52.70 &49.84\\
\gr AT+AWP+WOT-B ({Ours})  &\textbf{59.26} &55.69 &\textbf{55.09} &\textbf{52.82}&\textbf{50.00}\\
\bottomrule
\end{tabular}
\end{sc}
\end{center}
\label{tab:atvariants}
\end{table*}

\subsection{WOT: In-Time Refining Optimization Trajectories}
WOT reconstructs a set of historical optimization trajectories in time during the course of training on an unseen set to avoid overfitting. A naive strategy is treating each optimization trajectory $w^{i}$ as a whole and learning an individual weight for each trajectory shown as Eq.~\ref{newdirection}. This simple method naturally has limited  learning space for refinement especially when we only have a few historical trajectories. To improve the learning space of WOT, we further propose blockwise WOT which breaks down each trajectory into multiple blocks based on the original block design of the model itself. For convenience, we dot these two methods WOT-W and WOT-B, respectively.

\textbf{WOT-W} takes one trajectory as whole and assigns a single $\alpha$ for each trajectory. Hence the number of $\alpha$ that need to be learned exactly equals to the number of Gaps:$k$. 

\textbf{WOT-B} in contrast learns a vector of $\alpha$\ whose  length is determined by the number of the model's block. Therefore, Eq.~\ref{newdirection} can be reformulated as:
\begin{small}
\begin{align}
    \widetilde{\Delta w}=
    \begin{bmatrix}
    \widetilde{\Delta w}_{1}\\
    ...\\
    \widetilde{\Delta w}_{j}\\
    ...\\
    \widetilde{\Delta w}_{t}
    \end{bmatrix},\;\;
    \widetilde{\Delta w}_{j}=\alpha^{1}_{j}\Delta w^{1}_{j} +\alpha^{2}_{j}\Delta w^{2}_{j}+...+\alpha^{k}_{j}\Delta w^{k}_{j}
\end{align}
\end{small}
where $j$ denotes the $j$-th block. Optimizing $\alpha$ for blockwise WOT is exactly the same as the description in Eq.~\ref{update_alpha_m} and Eq.~\ref{update_alpha}.

The main difference between WOT-W and WOT-B is that WOT-W learns one $\alpha$ for each cached $\Delta w$ (the difference of parameters in two checkpoints) whereas WOT-B further breakdowns each cached $\Delta w$ into several blocks (each block contains several layers depending on the specific architectures as explained in detail in Appendix~B) and learns an individual $\alpha$ value for each block. Therefore, the learning space of WOT-B is larger than WOT-W, leading to better performance in general. The pseudocode of WOT can be found in Appendix~C. 
\section{Experiments}
\label{exp}
We perform extensive experiments to show the effectiveness of our method in improving adversarial robustness as well as addressing the robust overfitting issue. 
\begin{table*}[tbh]
\caption{Test robustness under multiple adversarial attacks based on  VGG-16/WRN-34-10 architectures. The experiments are conducted on CIFAR-10 with AT and Trades. The bold denotes the best performance.}
\begin{center}
\begin{sc}
\begin{adjustbox}{width=0.9\textwidth}
\begin{tabular}{c|c|ccccc}
\toprule
\multirow{1}{*}{Architecture} & \multirow{1}{*}{Method} &CW$_{\infty}$&PGD-20 & PGD-100&AA-$L_{\infty}$ \\
\midrule
\multirow{1}{*}{VGG16} &AT+early stop&46.87&49.95&46.87&43.63\\
VGG16&\tabincell{c}{AT+SWA}& 47.01&49.58&49.13&43.89\\
\gr VGG16 &\tabincell{c}{AT+WOT-W(Ours)}&47.42&49.96&49.36&44.01\\
\gr VGG16&\tabincell{c}{AT+WOT-B(Ours)}&\textbf{47.52}&\textbf{50.28}&\textbf{49.58}&\textbf{44.10}\\
\cline{1-6}
VGG16&TRADES&45.47&48.24&47.54&43.64\\
 VGG16&\tabincell{c}{TRADES+SWA}&45.92&48.64&47.86&44.12\\
\gr VGG16 &\tabincell{c}{TRADES+WOT-W(Ours)}&\textbf{46.75}&\textbf{49.19}&\textbf{48.28}&\textbf{44.82}\\
\gr VGG16 &\tabincell{c}{TRADES+WOT-B(Ours)}&46.21&48.81&47.85&44.17\\
\midrule
\midrule
\multirow{1}{*}{WRN-34-10} &AT+early stop&53.82&55.06&53.96&51.77\\
 WRN-34-10&\tabincell{c}{AT+SWA}&56.04&55.34&53.61&52.25\\
\gr WRN-34-10 &\tabincell{c}{AT+WOT-W(Ours)}&56.05&58.21&57.11&52.88\\
\gr WRN-34-10 &\tabincell{c}{AT+WOT-B(Ours)}&\textbf{57.13}&\textbf{60.15}&\textbf{59.38}&\textbf{53.89}\\
\cline{1-6}
WRN-34-10&TRADES&54.20&56.33&56.07&53.08\\
WRN-34-10 &\tabincell{c}{TRADES+SWA}&54.55&54.95&53.08&51.43\\
\gr WRN-34-10 &\tabincell{c}{TRADES+WOT-W(Ours)}&56.10&57.56&56.20&53.68\\
\gr WRN-34-10 &\tabincell{c}{TRADES+WOT-B(Ours)}&\textbf{56.62}&\textbf{57.92}&\textbf{56.80}&\textbf{54.33}\\
\bottomrule
\end{tabular}
\end{adjustbox}
\end{sc}
\end{center}
\label{tab:archi}
\vskip -0.1in
\end{table*}

\textbf{Datasets.} Four datasets are considered in our experiments: CIFAR-10, CIFAR-100~\cite{krizhevsky2010cifar}, Tiny-ImageNet~\cite{deng2009imagenet} and SVHN~\cite{netzer2011reading}. For experiments of WOT, we randomly split 1000 samples from the original CIFAR-10 training set, 10000 samples from Tiny-ImageNet, and 2000 samples from the original CIFAR-100 and SVHN training set as the unseen hold-out sets.

\textbf{Baselines.} Five baselines are included: AT~\cite{rice2020overfitting}, Trades~\cite{zhang2019theoretically}, AWP+AT~\cite{wu2020adversarial}, MART~\cite{wang2019improving} and SWA~\cite{chen2020robust}. Three architectures including VGG-16~\cite{simonyan2014very}, PreActResNet-18 (PreRN-18)~\cite{He2016}, WideResNet-34-10 (WRN-34-10)~\cite{zagoruyko2016wide}. 

\textbf{Experimental Setting.} 
 For WOT, we adopt an SGD optimizer with a momentum  of 0.9, weight decay of 5e-4, and a total epoch of 200 with a batch size of 128 following~\cite{rice2020overfitting}. By default, we start to refine optimization trajectories after 100 epochs. For WOT-B, we set each block in PreRN-18 and WRN-34-10 architectures as the independent weight space. We set the layers with the same width as a group and set each group as an independent block for VGG-16 (see details in Appendix~B). We by default set the gaps $m$ to 400, the number of gaps $k$ to 4 and initialize $\alpha$ as zero.  For all baselines, we use the training setups and hyperparameters exactly the same as their papers (see details in Appendix~A). 

\textbf{Evaluation Setting.}
We use AA attack~\cite{croce2020reliable} as our main adversarial robustness evaluation method. AA attack is a parameter-free ensembled adversarial attack that contains three white-box attacks:  APGD-CE~\cite{croce2020reliable}, APGD-T~\cite{croce2020reliable}, FAB-T~\cite{croce2020minimally} and one black-box attack: Square attack~\cite{andriushchenko2020square}. To the best of our knowledge, AA attack is currently the most reliable adversarial attack for evaluating adversarial robustness. We also adopt three other commonly used white-box adversarial attacks: FGSM~\cite{Goodfellow2014}, PGD-20/100~\cite{madry2017towards} and C\&W$_\infty$ attack~\cite{carlini2017towards}. Besides, we also report the performance of query-based SPSA black-box attack~\cite{uesato2018adversarial} (100 iterations with a learning rate of 0.01 and 256 samples for each gradient estimation). By default, we report \uline{the mean of three random runs} for all experiments of our method and omit the standard deviation since it is very small ($\leq 0.3\%$). We by default set $\epsilon=8/255$ for $L_{\infty}$ version adversarial attack and $\epsilon=64/255$ for $L_{2}$ version adversarial attack.

\begin{table*}[!h]
\caption{Test robustness under AA-$L_{2}$ and AA-$L_{\infty}$ attacks across various datasets. The experiments are based on PreRN-18 and AT. The bold denotes the best performance.}
\begin{center}
\begin{sc}
\begin{adjustbox}{width=0.8\textwidth}
\begin{tabular}{c|c|cccccccc}
\toprule
\multirow{1}{*}{Attack} & \multirow{1}{*}{Method} &\multicolumn{1}{c}{SVHN}& \multicolumn{1}{c}{CIFAR-10}&\multicolumn{1}{c}{CIFAR-100}&\multicolumn{1}{c}{Tiny-ImageNet}\\
\midrule
\multirow{1}{*}{L$_{\infty}$} &AT+early stop&45.72&47.43&23.69&14.39\\
L$_{\infty}$&AT+SWA&40.24&48.61&23.90&17.94\\
\gr L$_{\infty}$&AT+WOT-W(Ours)&50.42&48.36&24.41&17.10\\
\gr L$_{\infty}$&AT+WOT-B(Ours)&\textbf{51.83}&\textbf{48.96}&\textbf{25.26}&\textbf{18.77}\\
 \midrule
\multirow{1}{*}{L$_{2}$} & AT+early stop&72.13&71.30&42.75&36.61\\
L$_{2}$&AT+SWA&67.76&73.28&43.10&42.40\\
\gr L$_{2}$&AT+WOT-W(Ours)&72.75&73.20&\textbf{43.88}&42.43\\
\gr L$_{2}$&AT+WOT-B(Ours)&\textbf{72.80}&\textbf{73.39}&43.32&\textbf{42.54}\\
\bottomrule
\end{tabular}
\end{adjustbox}
\end{sc}
\end{center}
\label{tab:dataset}
\end{table*}

\begin{figure*}[htb]
\vskip -0.1 in
\centering
\begin{adjustbox}{width=0.9\textwidth}
\includegraphics[width=0.9\textwidth]{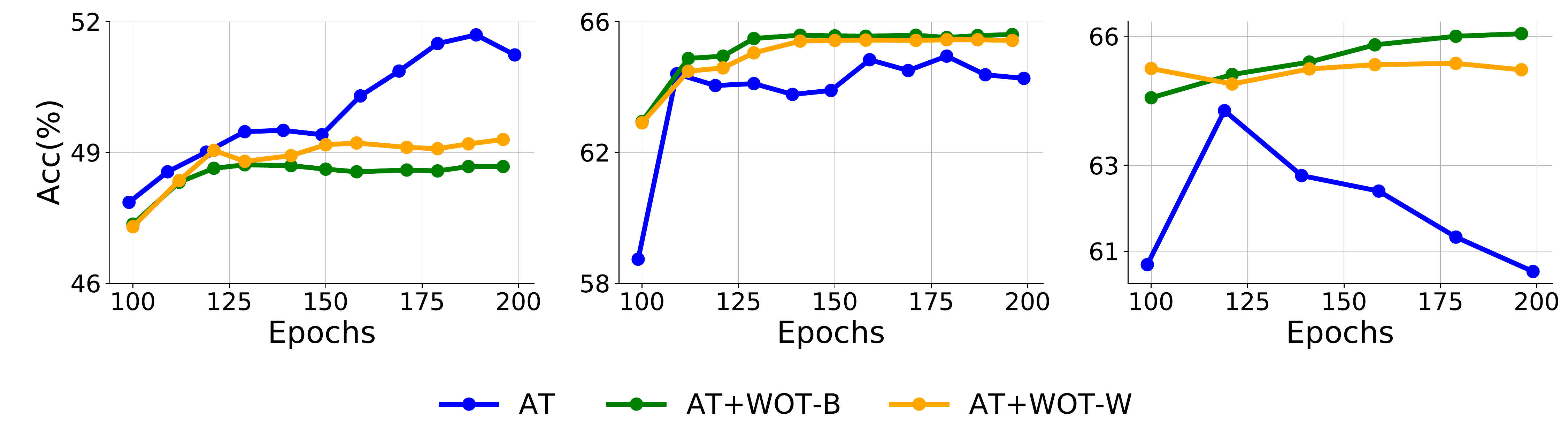}
\end{adjustbox}
\caption{Robust accuracy under black-box attack over epochs. \textbf{(Left)} Robust accuracy on the unseen robust model transfer attacked from checkpoints of AT, AT+WOT-W/B. \textbf{(Middle)} Robust accuracy on checkpoints of AT, AT+WOT-W/B  transfer attacked from the  unseen model. \textbf{(Right)} Robust accuracy on checkpoints of AT, AT+WOT-W/B under SPSA black-box attack. The experiments are conducted on PreRN-18 and CIFAR-10. The unseen robust model is WRN-34-10 trained by AT.}
\label{fig:transferattack}
\end{figure*}
\subsection{Superior Performance in Improving Adversarial Robustness} \label{superioriRA}
We evaluate the effectiveness of WOT in improving adversarial robustness across AT and three of its variants, four popular used datasets, i.e., SVHN, CIFAR-10, CIFAR-100 and Tiny-ImageNet, and three architectures, i.e., VGG16, PreRN-18, and WRN-34-10. 

\textbf{WOT consistently improves the adversarial robustness of all adversarial training variants.} In Table~\ref{tab:atvariants}, we applied WOT-B and WOT-W to AT+early stop, Trades, MART, and AWP variants and compare them with their counterpart baselines. Besides, we add the combination of SWA and these adversarial training variants as one of the baselines. The results show \textbf{\circled{1}} WOT consistently improves adversarial robustness among the four adversarial training variants under both weak attacks, e.g. FGSM, PGD-20, and strong attacks, e.g., C\&W$_\infty$, AA-$L_{\infty}$ attacks. \textbf{\circled{2}} WOT-B as the WOT variant confirms our hypothesis and consistently performs better than WOT-W. WOT-B improves the robust accuracy over their counterpart baselines by $0.35\% \sim 1.53\%$ under AA-$L_{\infty}$ attack. 
\textbf{\circled{3}} WOT boosts robust accuracy with a larger margin on AT and Trades than MART and AWP under AA-$L_{\infty}$ attack. One reason might be that MART and AWP themselves enjoy good ability in mitigating robust overfitting~\cite{stutz2021relating,wu2020adversarial}, leading to less space for WOT to further boost the performance. 

\textbf{WOT can generalize to different architectures and datasets.}  Table~\ref{tab:archi} and Table~\ref{tab:dataset} show that WOT consistently outperforms the counterpart baseline under AA-$L_{\infty}$ attack, which indicates that the effectiveness of WOT generalizes well to different architectures and datasets. In Table~\ref{tab:archi}, WOT boosts robust accuracy by $0.47\% \sim 2.12\%$ on VGG16 and WRN-34-10 architectures.  In Table~\ref{tab:dataset}, WOT improves robust accuracy with $1.53\% \sim 6.11\%$ among SVHN, CIFAR-10, CIFAR-100 and Tiny-ImageNet under AA-$L_{\infty}$ attack. Besides, the success of WOT can also be extended to AA-$L_{2}$ attack with the improvement by $0.67\% \sim 5.93\%$.

\begin{figure*}[tbh]
\centering
\includegraphics[width=0.9\textwidth]{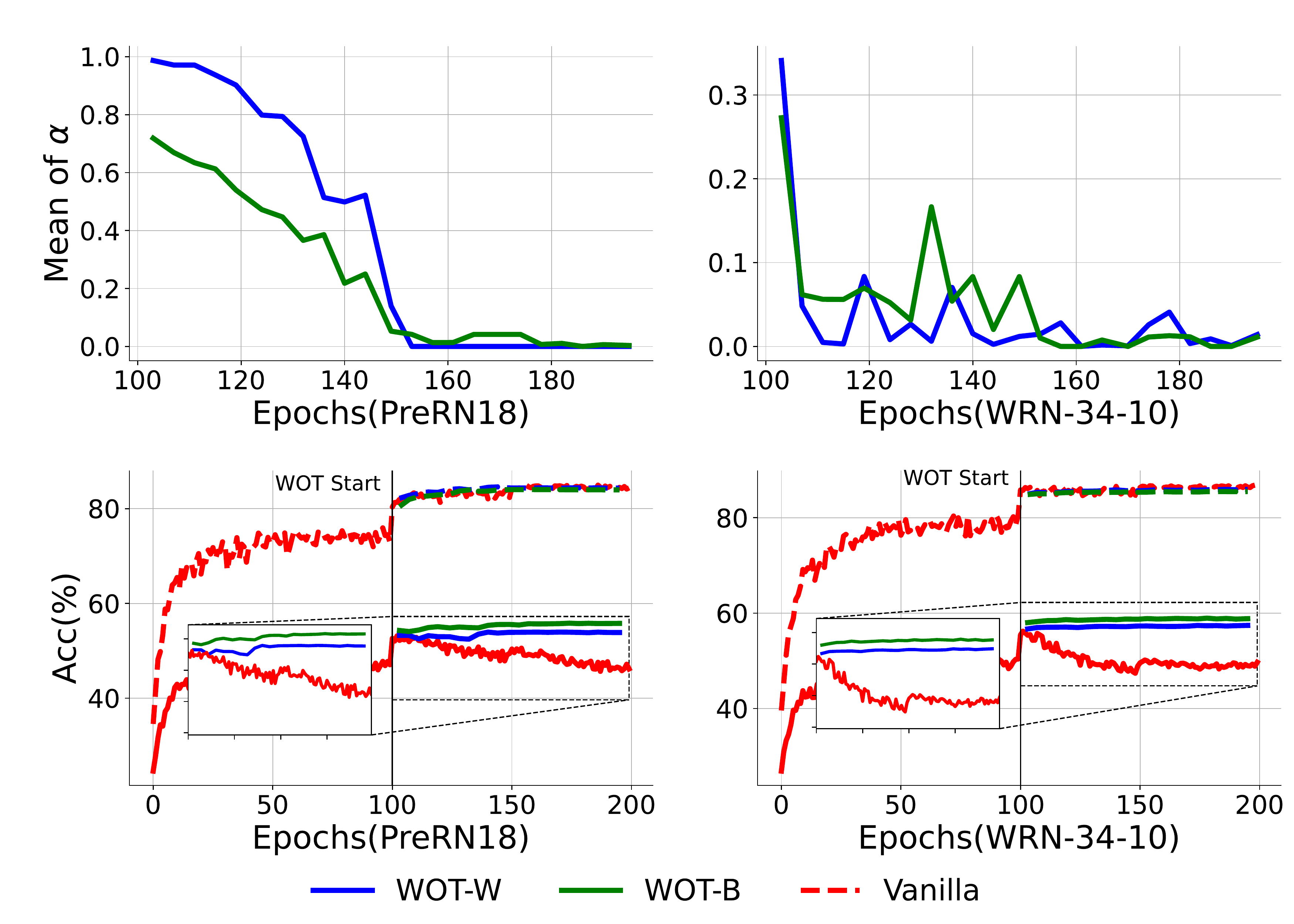}
\caption{Mean value of $\alpha$ and results of test robust/clean accuracy over epochs. The experiments are conducted on CIFAR-10 with PreRN-18 based on AT.}
\label{fig:overfitting}
\end{figure*}

\begin{table*}[tbh]
\caption{Test robustness under AA-$L_{\infty}$ attack to show the robust overfitting issue in AT and the effectiveness of WOT in overcoming it. The difference between the best and final checkpoints indicates performance degradation during training and the best checkpoint is chosen by PGD-10 attack on the validation set. The experiments are conducted on CIFAR-10 with PreRN-18/WRN-34-10 architectures.}
\begin{center}
\begin{sc}
\begin{adjustbox}{width=0.9\textwidth}
\begin{tabular}{c|c|ccc|ccc}
\toprule
\multirow{2}{*}{Architectures} & \multirow{2}{*}{Method} &\multicolumn{3}{c|}{Robust Accuracy(RA)}& \multicolumn{3}{c}{Standard Accuracy(SA)}\\
&&Best&Final &Diff. &Best &Final & Diff.\\
\midrule
\multirow{1}{*}{PreRN-18} &AT&48.02&42.48&-5.54&81.33&84.40&+3.07\\
 PreRN-18&AT+SWA&48.93&48.61&-0.32&84.19&85.23&+1.04\\
\gr PreRN-18&AT+WOT-W(Ours)&48.04&48.36&+0.32&84.05&84.47&-0.42\\
\gr PreRN-18&AT+WOT-B(Ours)&48.90&48.96&+0.06&83.84&83.83&-0.01\\
 \midrule
\multirow{1}{*}{WRN-34-10} &AT&51.77&46.78&-4.99&85.74&86.34&+0.6\\
 WRN-34-10&AT+SWA&53.38&52.25&-1.13&87.14&88.45&+1.31\\
\gr WRN-34-10&AT+WOT-W(Ours)&52.84&52.88&+0.04&84.83&84.88&+0.05\\
\gr WRN-34-10&AT+WOT-B(Ours)&52.23&53.89&+1.66&83.46&85.50&+2.04\\
\bottomrule
\end{tabular}
\end{adjustbox}
\end{sc}
\end{center}
\label{tab:overfitting}
\vskip -0.1 in
\end{table*}

\textbf{Excluding Obfuscated Gradients.} \cite{athalye2018obfuscated} claims that obfuscated gradients can also lead to the ``counterfeit" of improved robust accuracy under gradients-based white-box attacks. To exclude this possibility, we report the performance of different checkpoints under transfer attack and SPSA black-box attack over epochs. In Figure~\ref{fig:transferattack}, the left figure shows robust accuracy of the unseen robust model on the adversarial examples generated by the PreRN-18 model trained by AT, AT+WOT-B, AT+WOT-W respectively with PGD-10 attack on CIFAR-10. A higher robust accuracy  on the unseen robust model corresponds to a weaker attack.  It can be seen that both AT+WOT-B and AT+WOT-W generate more transferable adversarial examples than AT.  Similarly, the middle figure shows the robust accuracy of the PreRN-18 model trained by AT, AT+WOT-B, AT+WOT-W on the adversarial examples generated by the unseen robust model. It can be seen that AT+WOT-B and AT+WOT-W can better defend the adversarial examples from the unseen model. What's more, in the right figure, we observe again that both AT+WOT-B and AT+WOT-W outperform AT under SPSA black-box attack over different checkpoints during training. All these empirical results sufficiently suggest that the improved robust accuracy of WOT is not caused by obfuscated gradients. 

\subsection{Ability to Prevent Robust Overfitting}
We report the robust accuracy under AA-$L_{\infty}$ attack for the best checkpoint and the last checkpoint based on PreRN-18 and WRN-34-10 architectures on CIFAR-10 (Table~\ref{tab:overfitting}). Besides, we show the robust accuracy curve under PGD-10 attack on different checkpoints over epochs (Figure~\ref{fig:overfitting}).   

In Figure~\ref{fig:overfitting}, the third and fourth figures show that after the first learning rate decay (at 100 epoch), there is a large robust accuracy drop for AT between the best checkpoint and the last checkpoint on both PreRN-18 and WRN-34-10 architectures. In comparison, there is completely no robust accuracy drop for AT+WOT-W/B between the best checkpoint and the last checkpoint on both PreRN-18 and WRN-34-10 architectures. In Table~\ref{tab:overfitting}, we further show the evidence that there is no robust accuracy drop for AT+WOT-B/W under stronger attack, i.e., AA-$L_{\infty}$ attack. From the first and second figures of Figure~\ref{fig:overfitting}, we observe that the mean of $\alpha$ decreases to a very small value after 150,100 epochs for PreRN-18 and WRN-34-10 respectively. The small mean of $\alpha$ indicates that  WOT stops the model's weights from updating with unexpected magnitudes, which prevents the occurrence of robust overfitting.

\begin{figure}[tbh]
\centering
\includegraphics[width=0.45\textwidth]{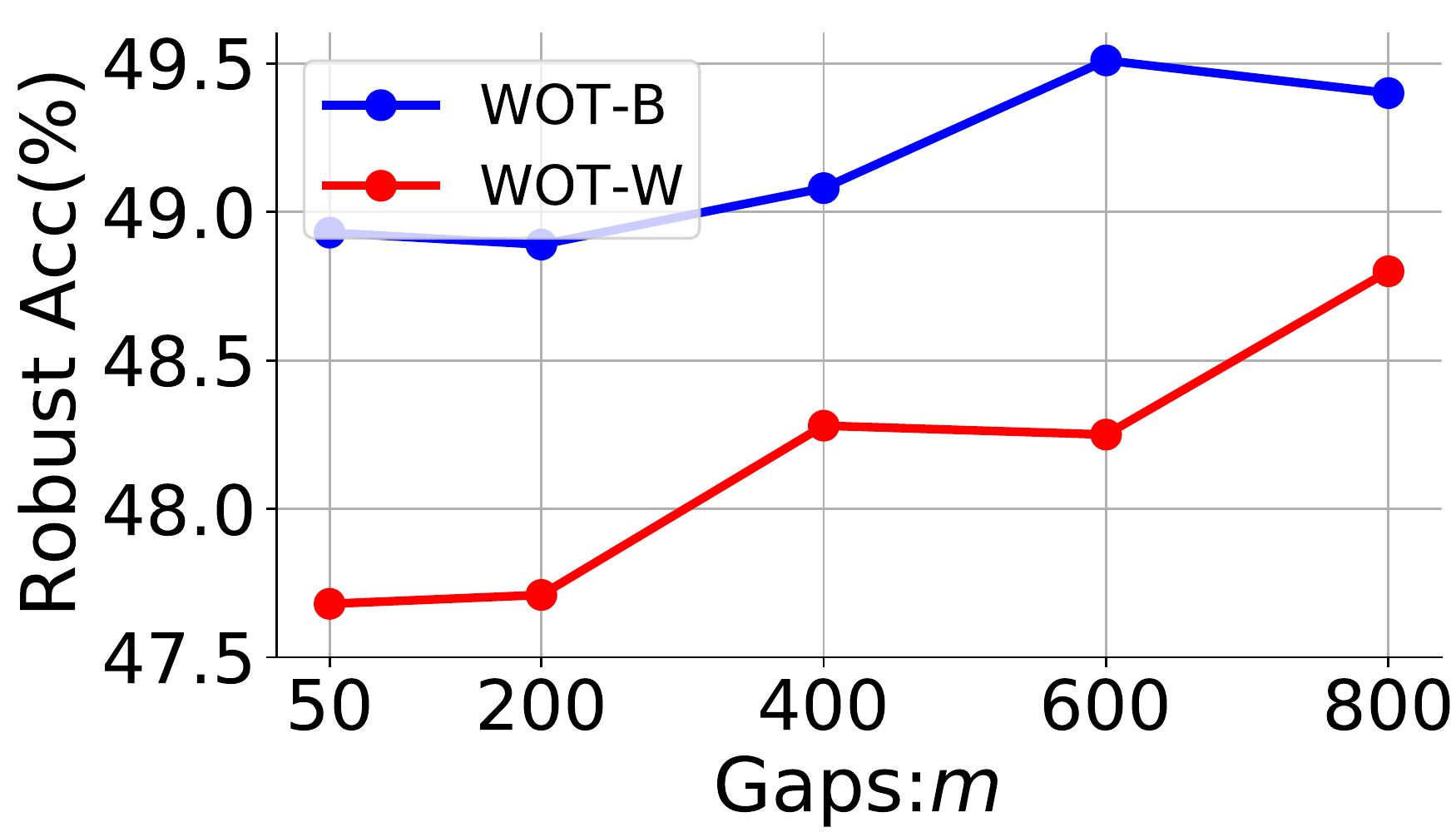}
\includegraphics[width=0.45\textwidth]{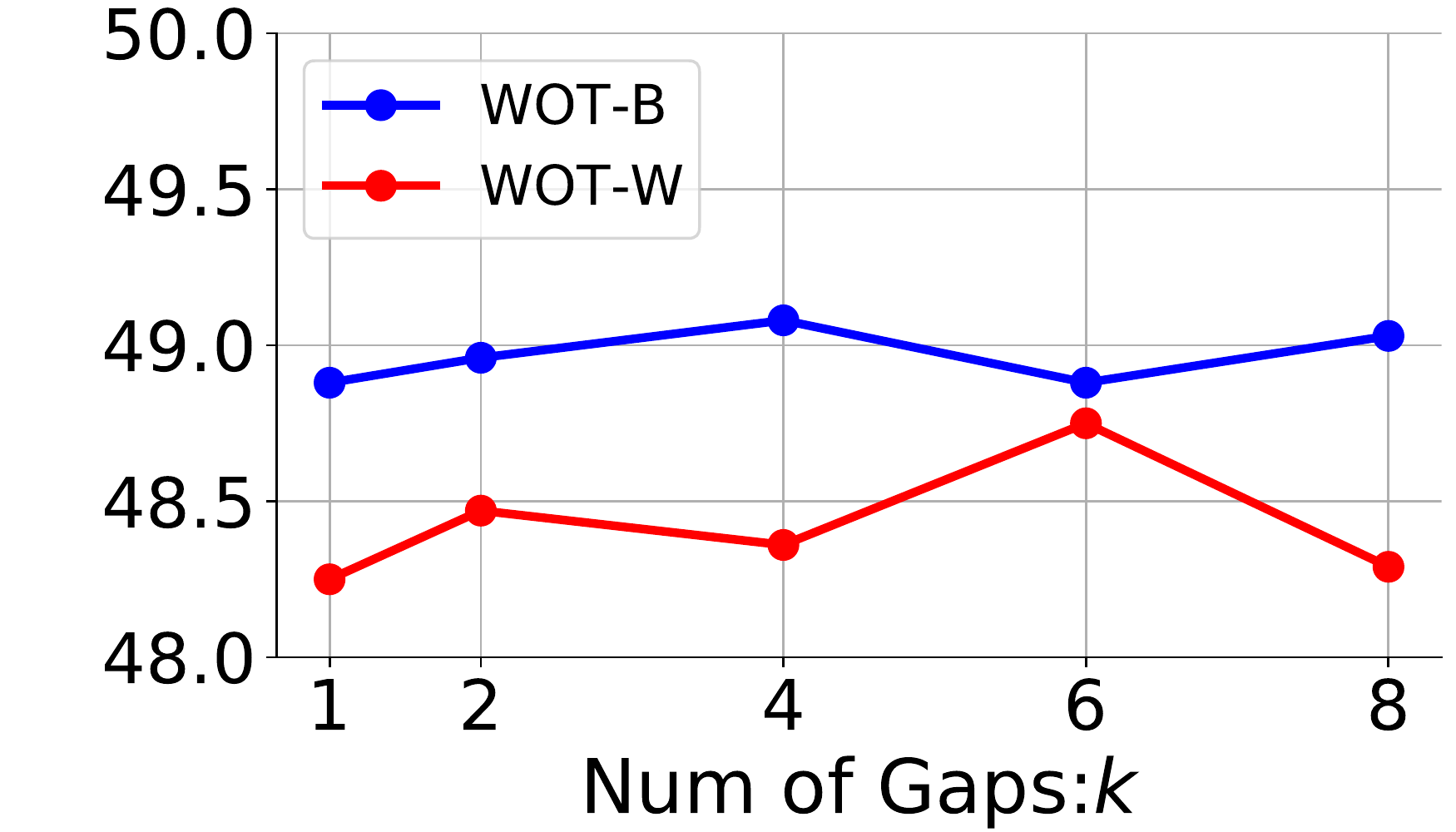}
\caption{The impact of gaps $m$ and the number of gaps $k$ on robust accuracy under AA-$L_{\infty}$ attack. The experiments are conducted on CIFAR-10 with PreRN-18 based on AT. $k$ is fixed to 4 for the left figure and $m$ is fixed to 400 for the right figure.}
\label{fig:impactofhyper}
\vskip -0.1 in
\end{figure}

\begin{figure*}[tbh]
\centering
\subfloat[Trades]{\includegraphics[width=0.45\textwidth]{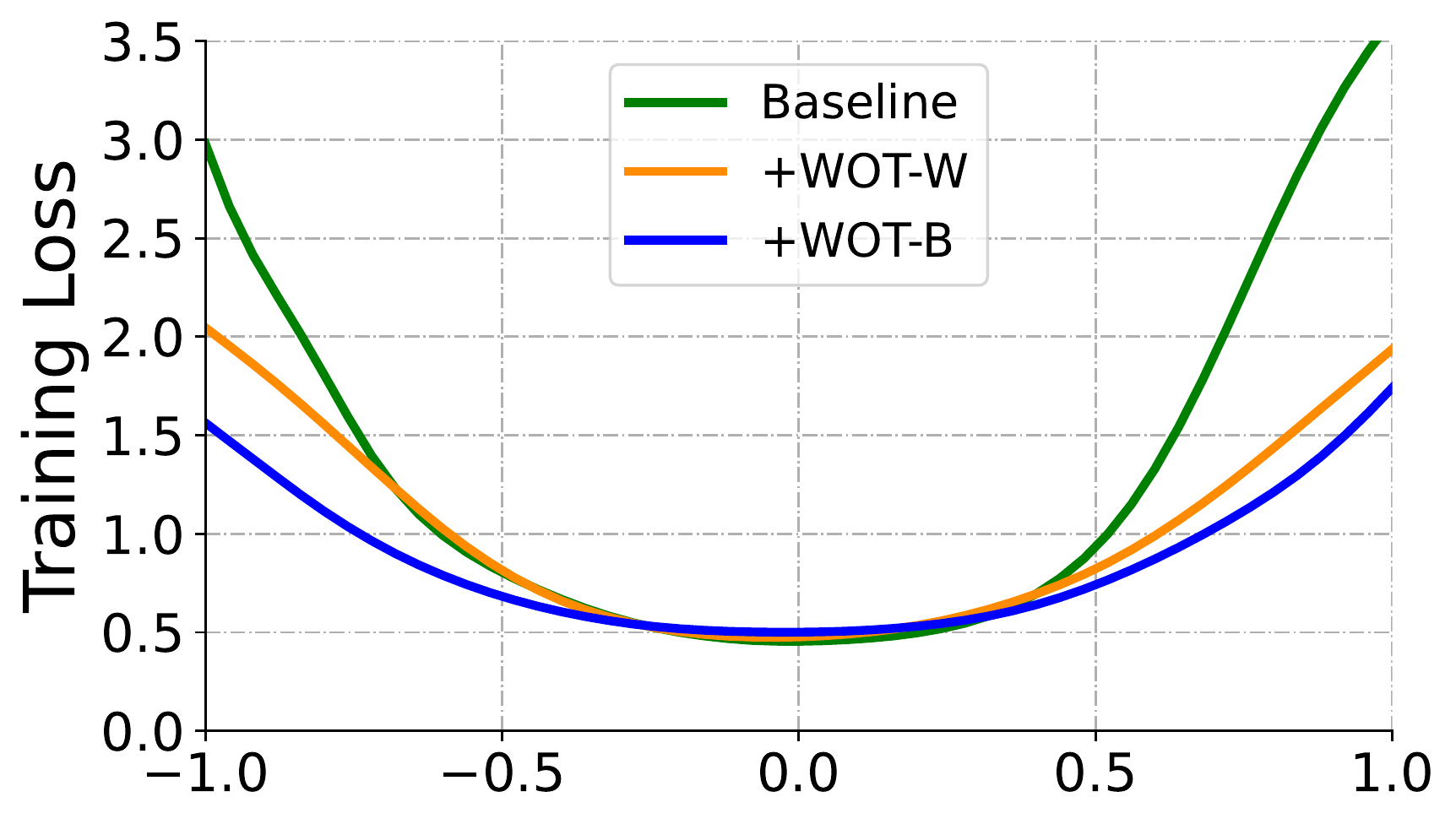}\label{fig:landscapes_weights:a}}
\subfloat[AT]{\includegraphics[width=0.45\textwidth]{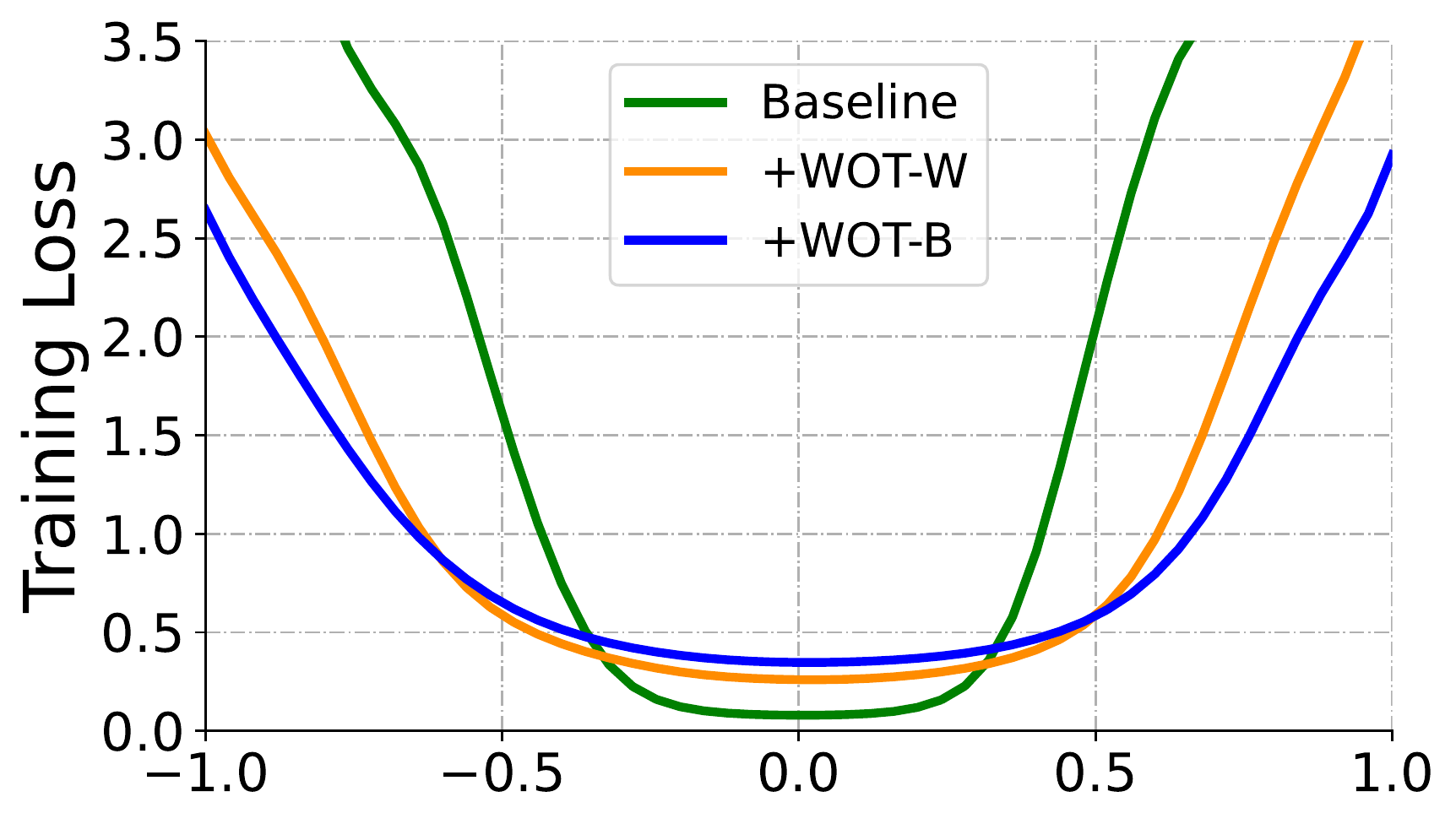}\label{fig:landscapes_weights:b}}
\\
\subfloat[Blocks (X-axis)]{\includegraphics[width=0.45\textwidth]{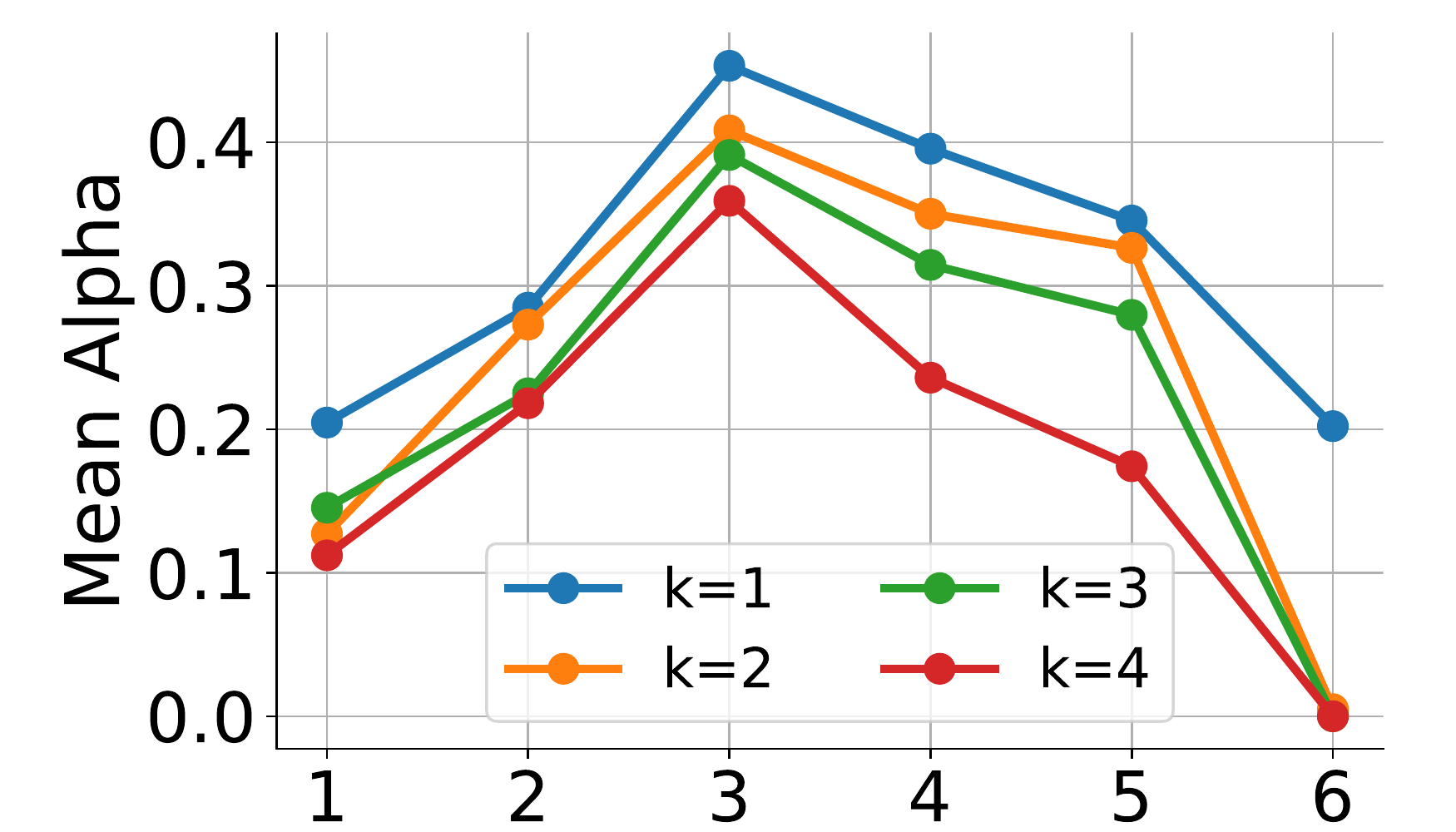}\label{fig:landscapes_weights:c}}
\subfloat[Epochs (X-axis)]{\includegraphics[width=0.45\textwidth]{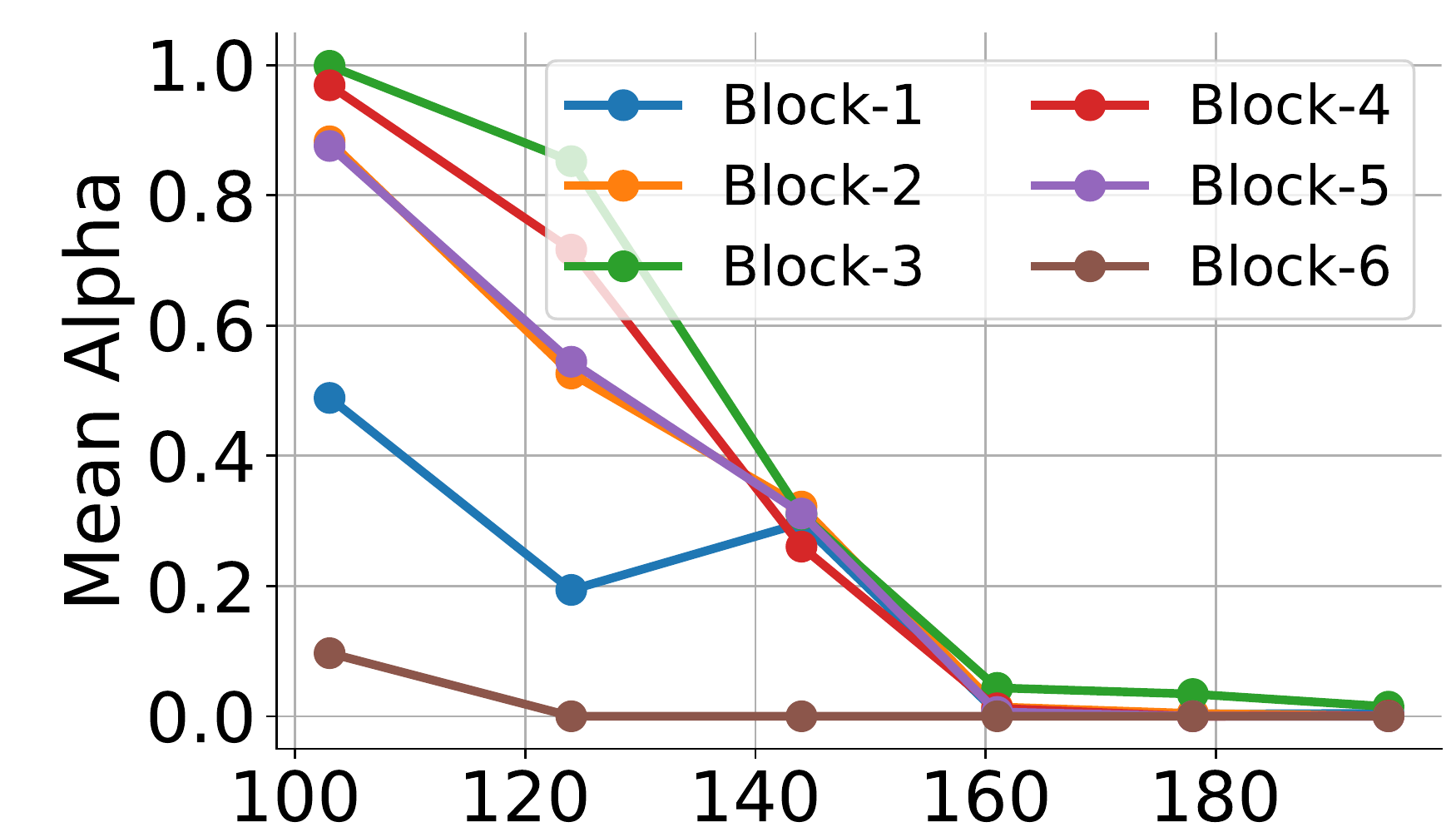}\label{fig:landscapes_weights:d}}
\caption{Loss landscape w.r.t weight space (Figure~\ref{fig:landscapes_weights:a} and Figure~\ref{fig:landscapes_weights:b}). z-axis denotes the loss value. We plot the loss landscape following the setting in~\cite{wu2020adversarial}. The averaged $\alpha$ by averaging along training process (Figure~\ref{fig:landscapes_weights:c}). The k-averaged $\alpha$ during the training process. (Figure~\ref{fig:landscapes_weights:d}). The experiments are conducted on CIFAR-10 with PreRN-18.}
\label{fig:landscapes_weights}
\vskip -0.1 in
\end{figure*}

\begin{table}[tbh]
\caption{Robust Accuracy of ablation experiments on CIFAR-10 with PreRN-18.}
\begin{center}
\begin{small}
\begin{sc}
\begin{adjustbox}{width=0.9\textwidth}
    \begin{tabular}{c|c|c|c|c}
    \toprule
        Methods  &PGD-20 &PGD-100&CW$_\infty$&AA-L$_\infty$\\
         \midrule
         AT+B1& 49.68&47.44&49.04&45.26\\
         AT+B2&52.74 &51.28 &51.31&48.22\\
         AT+WOT-B+B3 (m=400,k=4)& 47.14&44.23&43.87&41.02\\
         \gr AT+WOT-W (m=400,k=4)&53.19& 51.90 &51.74 &48.36\\
        \gr  AT+WOT-B (m=400,k=4)&54.85&53.77&52.56&48.96\\
         \bottomrule
    \end{tabular}
    \end{adjustbox}
        \end{sc}
\end{small}
\end{center}
    \label{tab:extrabaslines_ciar10}
\end{table}

 \begin{figure*}[tbh]
 \centering
\setlength\tabcolsep{1pt}
\settowidth\rotheadsize{Radcliffe Cam}
\begin{tabularx}{0.9\linewidth}{l XXXX }
\rothead{\centering{AT}} 
                        &   \includegraphics[width=\hsize,valign=m]{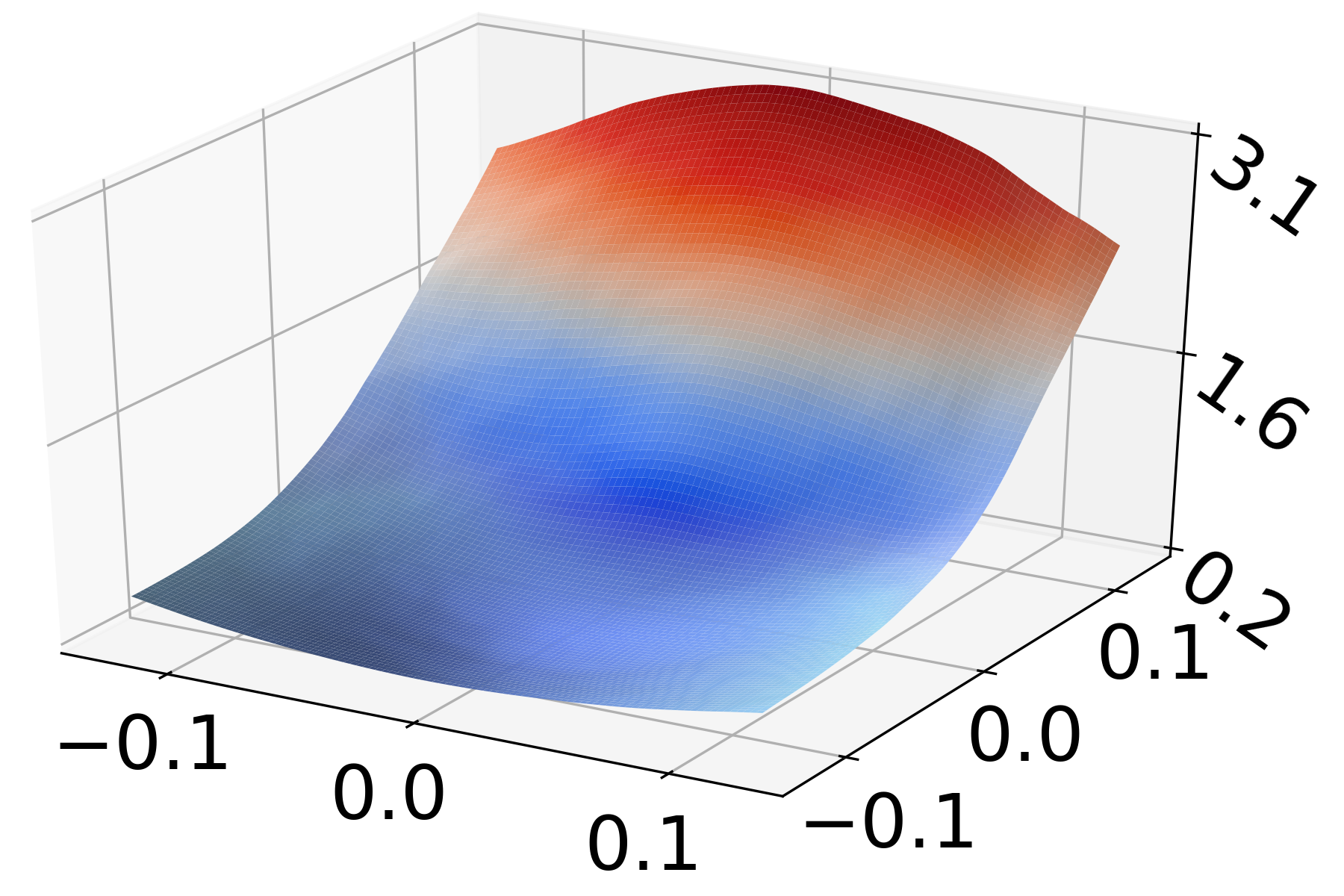}    
                        &   \includegraphics[width=\hsize,valign=m]{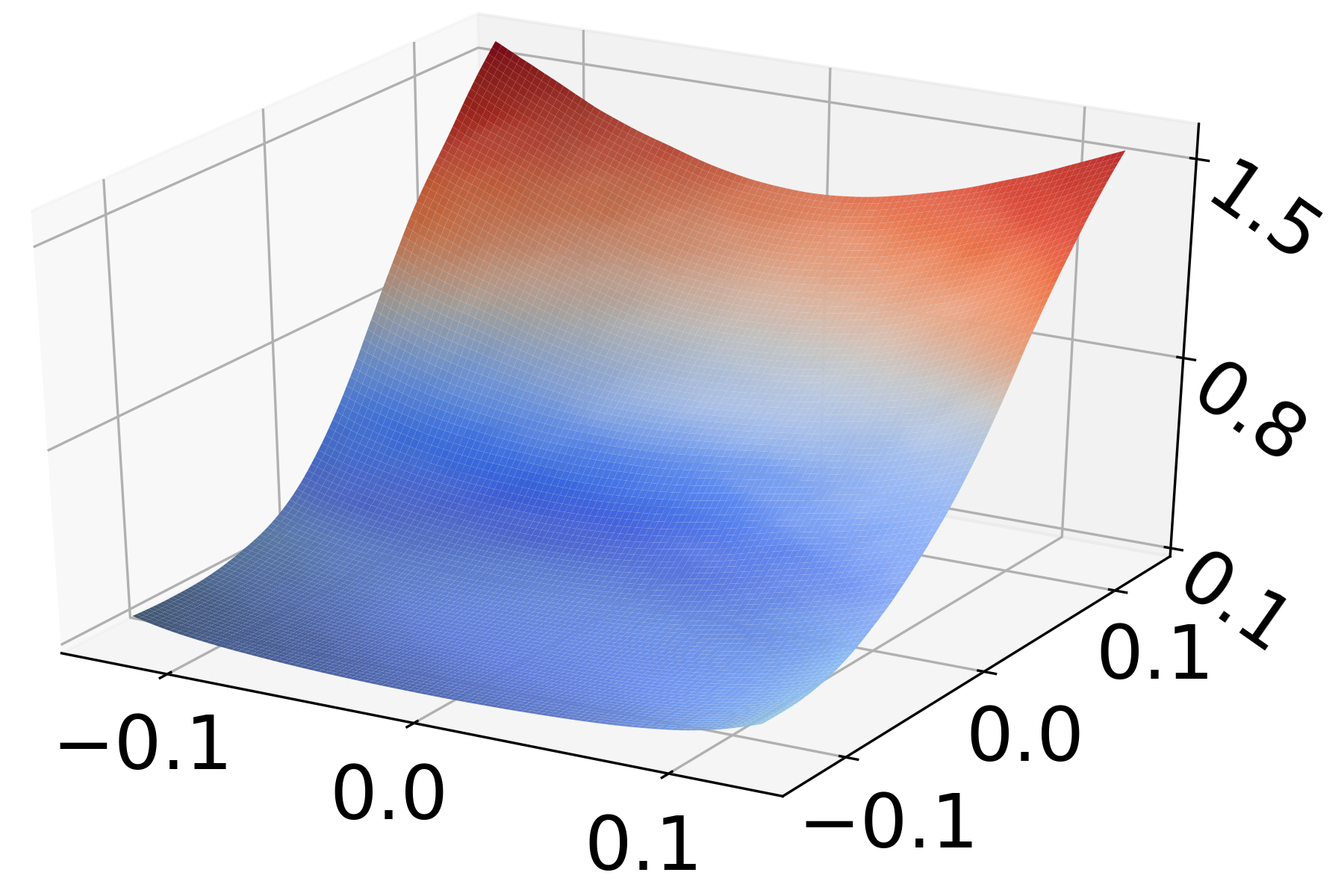}
                        &   \includegraphics[width=\hsize,valign=m]{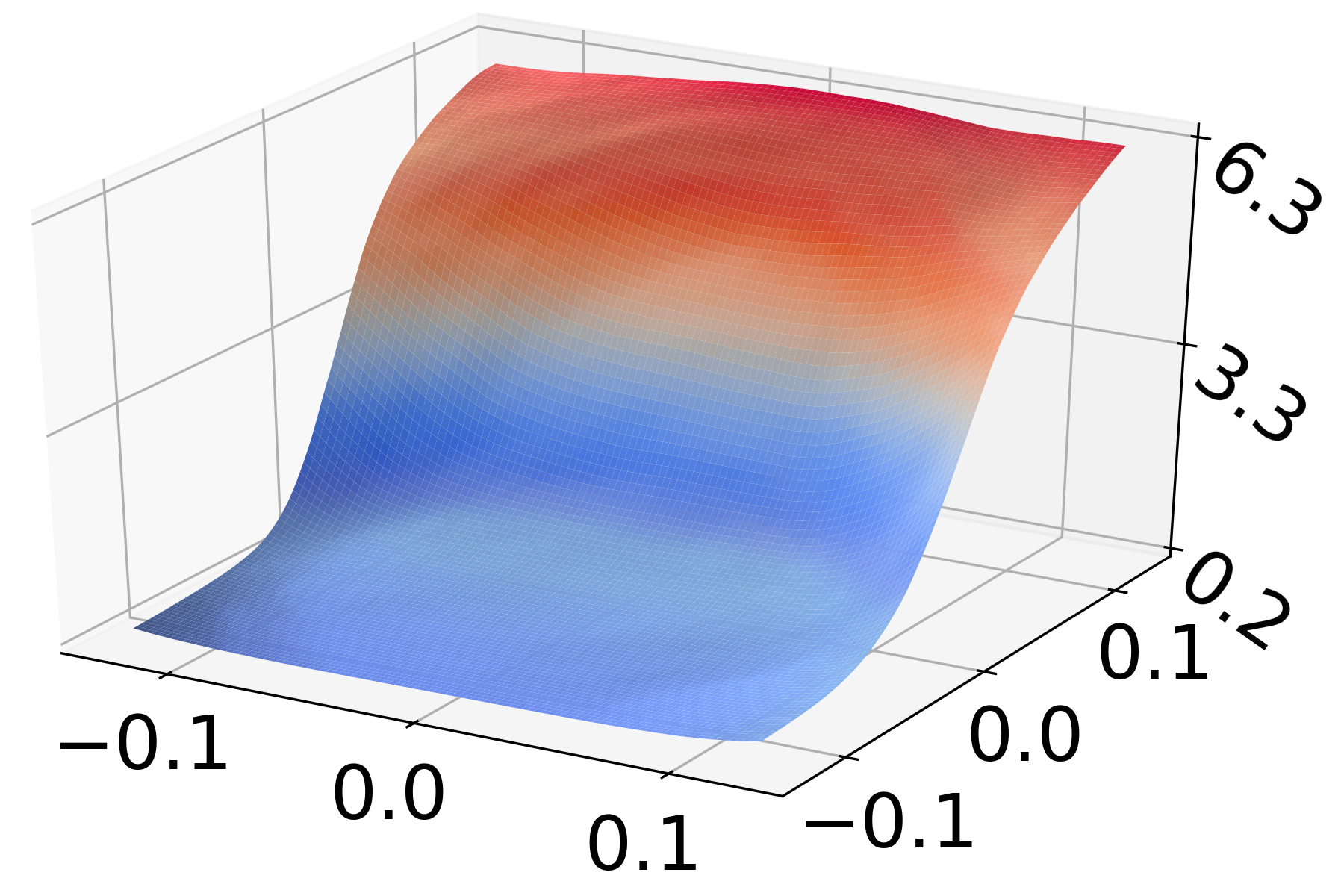}    
                        &   \includegraphics[width=\hsize,valign=m]{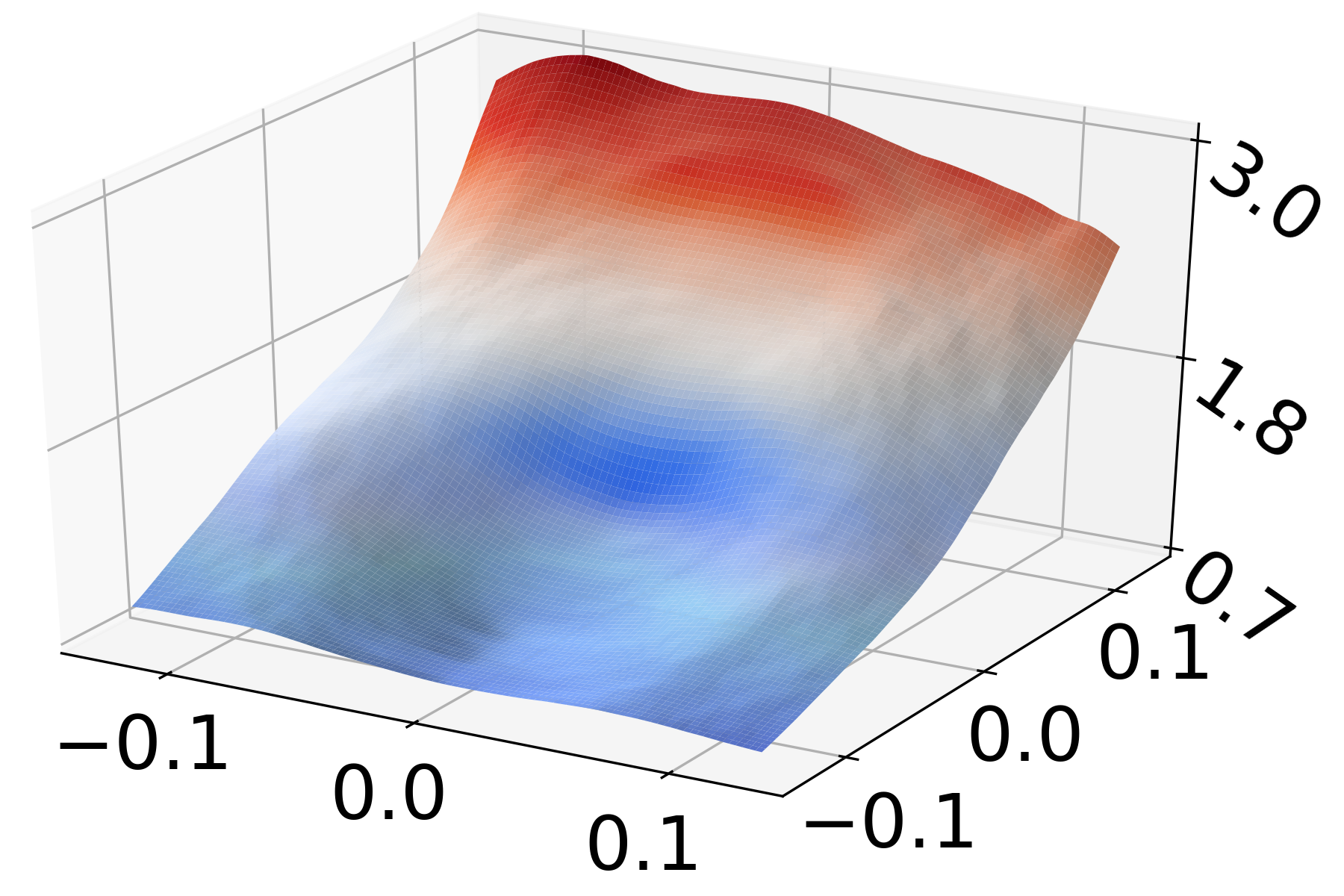}\\
\rothead{\centering{AT+WOT-W}} 
                        &   \includegraphics[width=\hsize,valign=m]{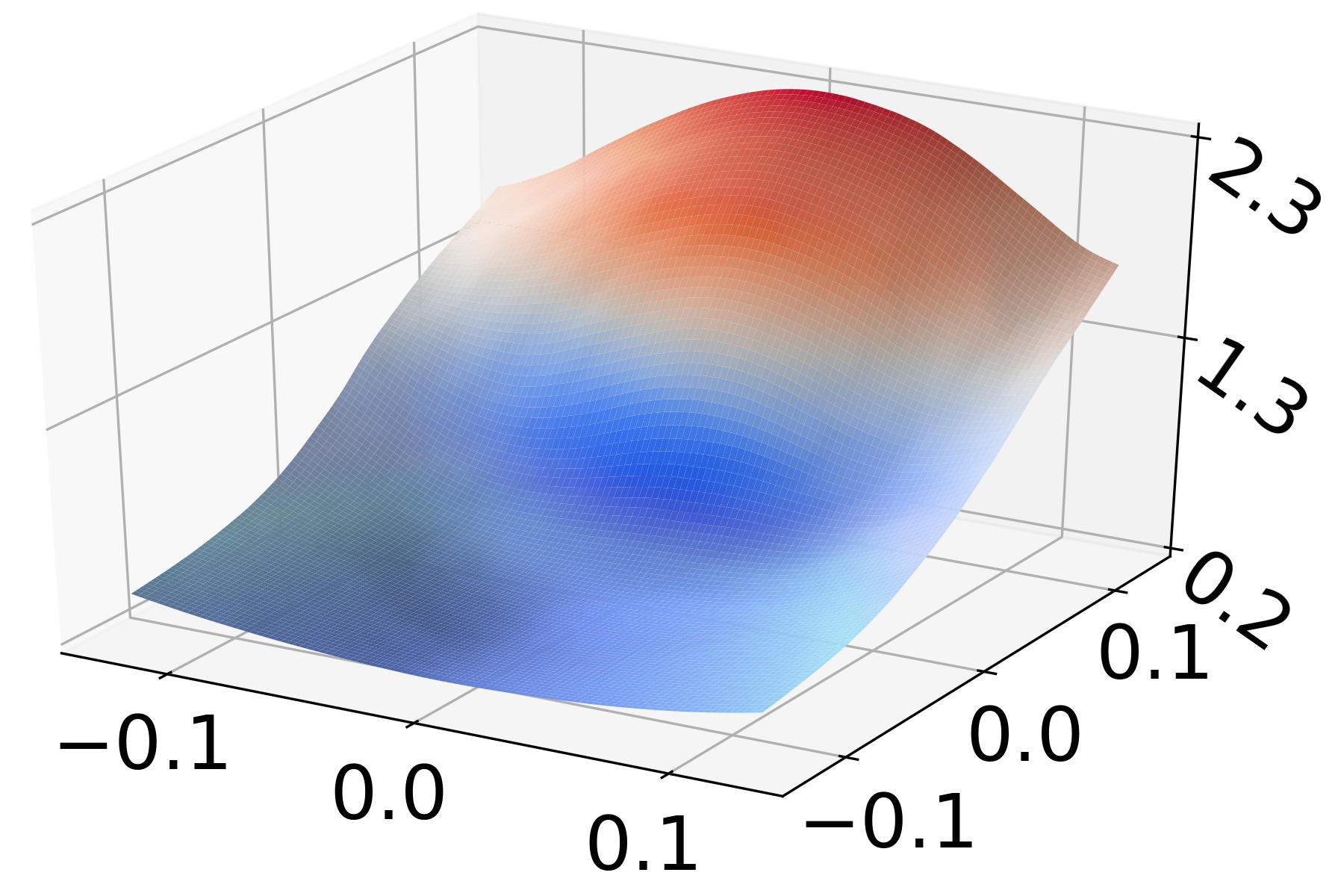}
                        &   \includegraphics[width=\hsize,valign=m]{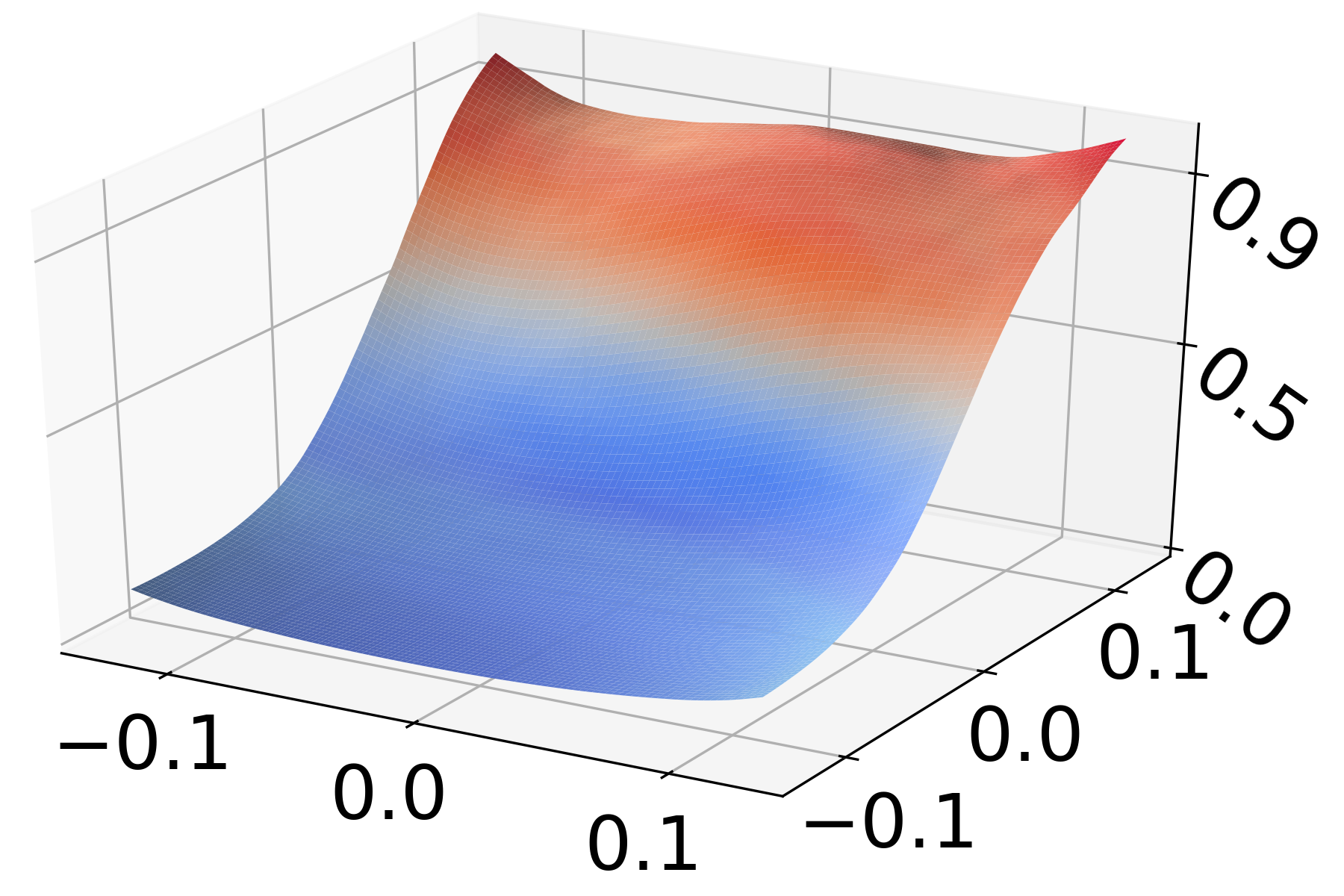}
                        &   \includegraphics[width=\hsize,valign=m]{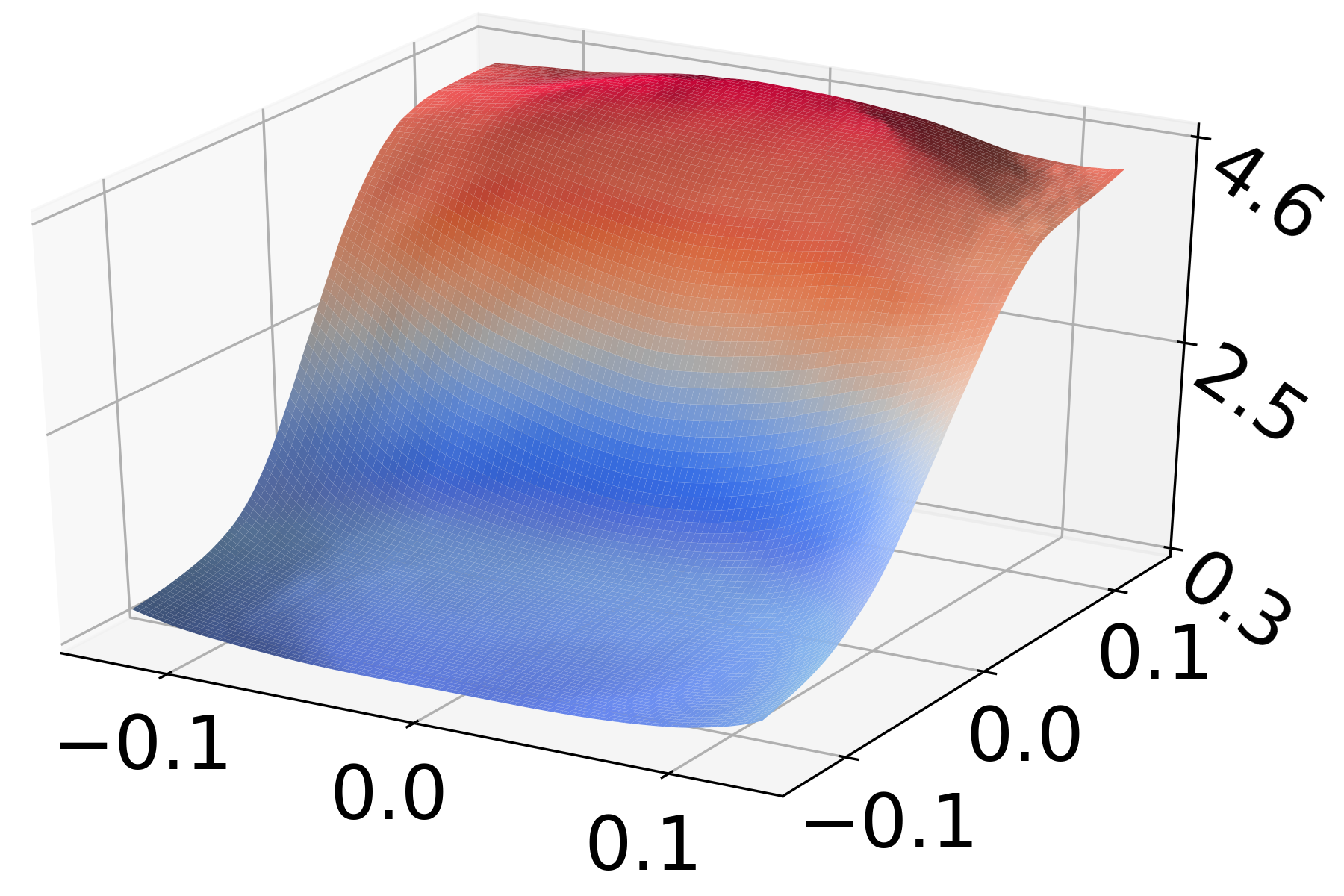}
                        &   \includegraphics[width=\hsize,valign=m]{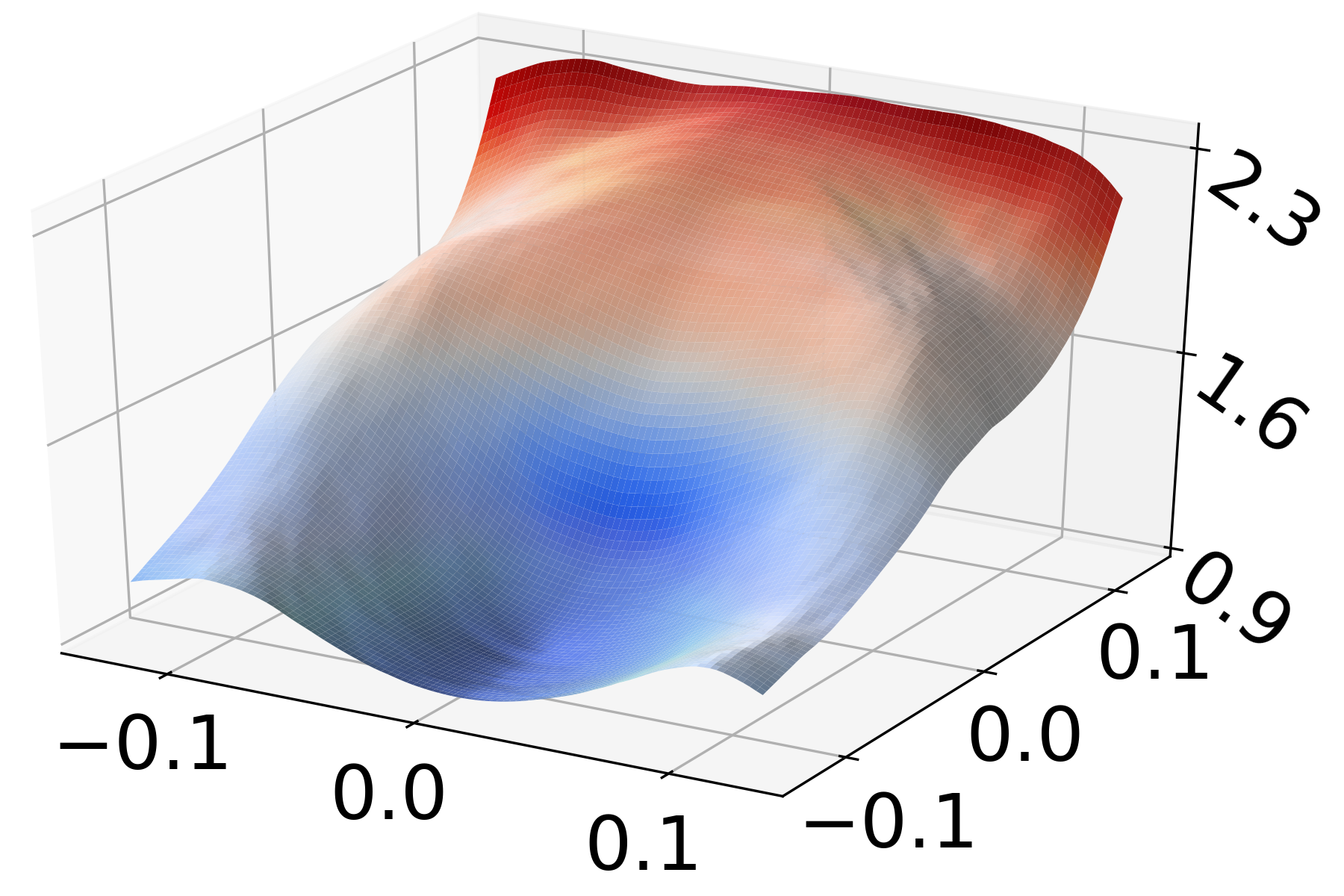}\\
\rothead{\centering{AT+WOT-B}} 
                        &   \includegraphics[width=\hsize,valign=m]{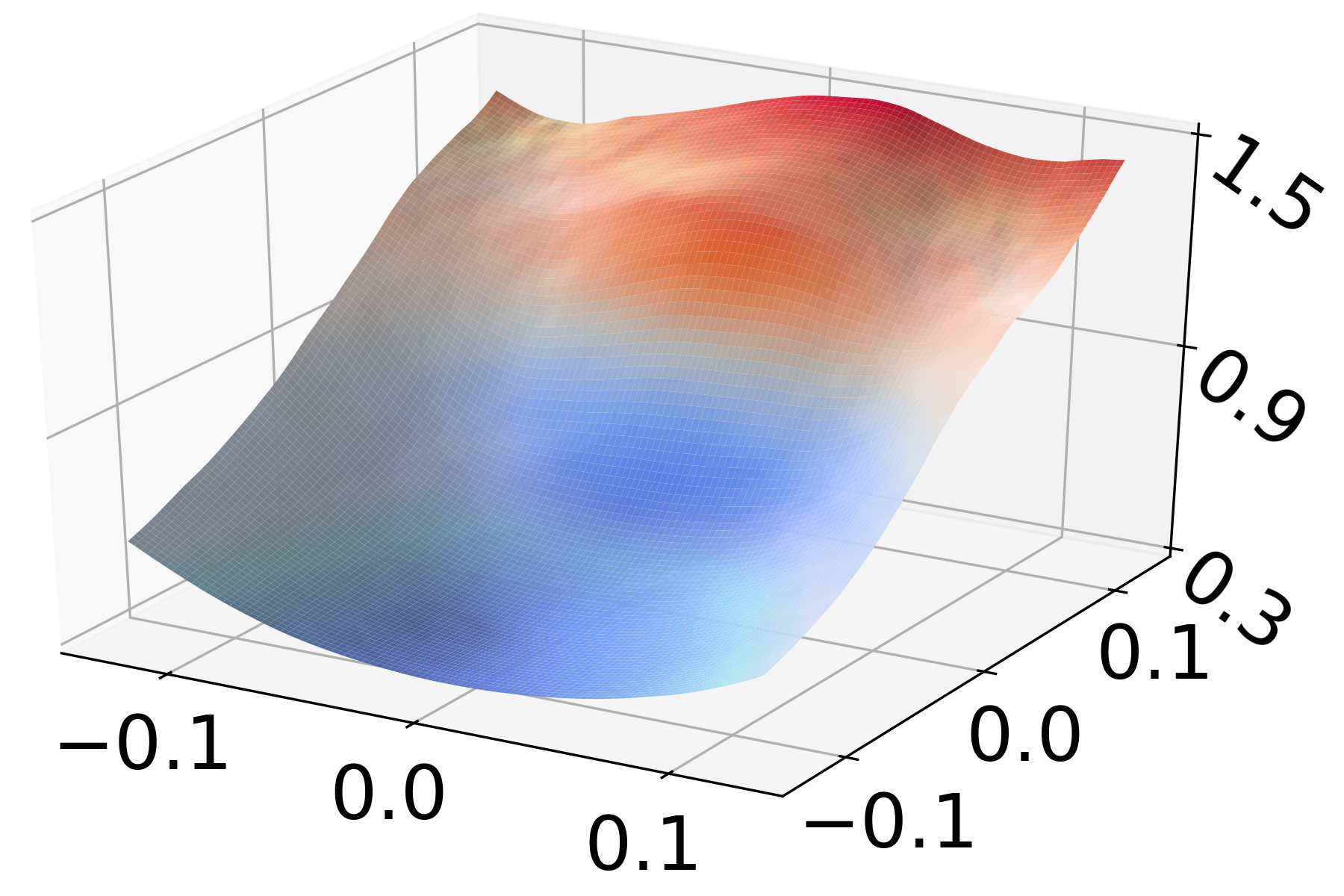}    
                        &   \includegraphics[width=\hsize,valign=m]{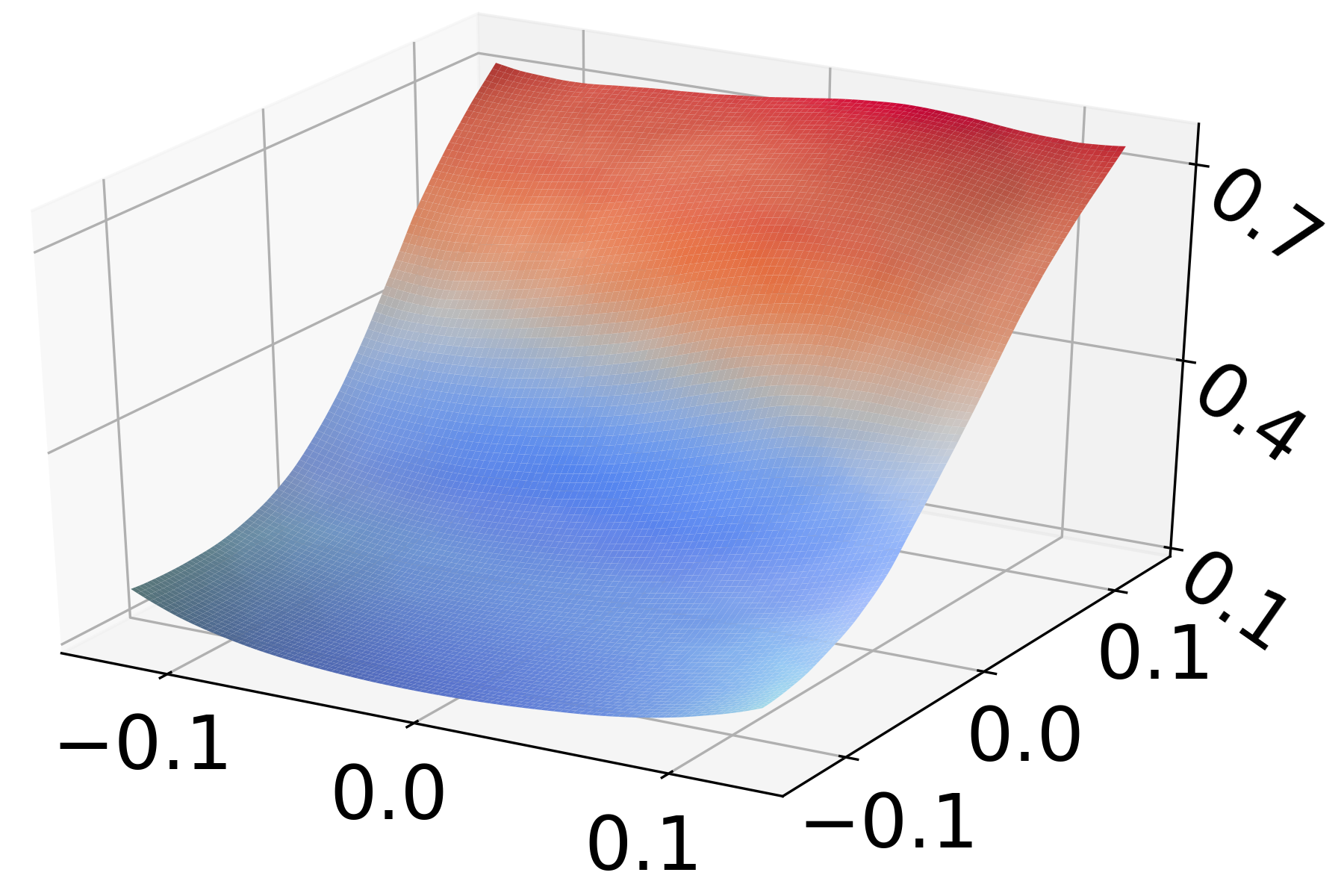}
                        &   \includegraphics[width=\hsize,valign=m]{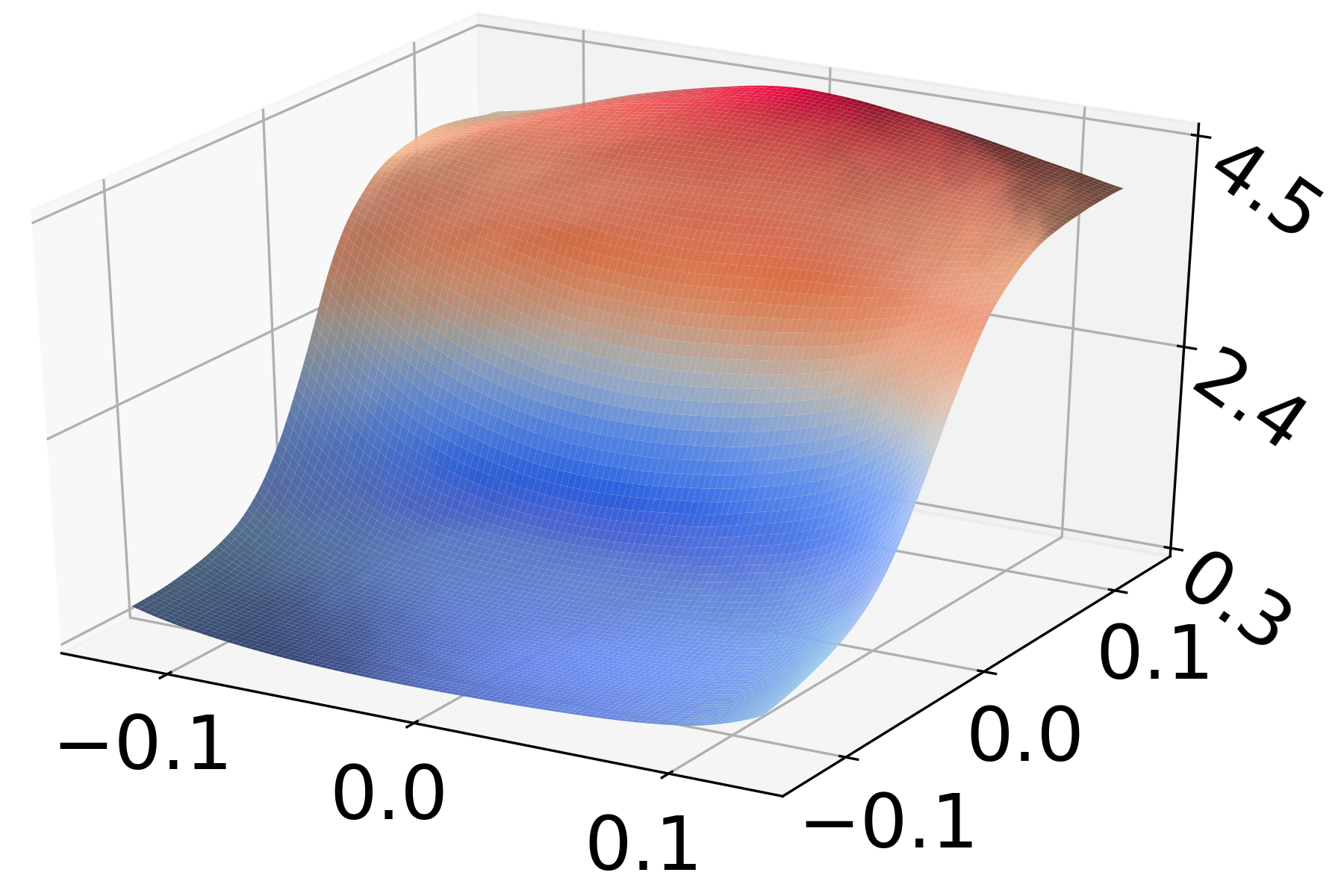}    
                        &   \includegraphics[width=\hsize,valign=m]{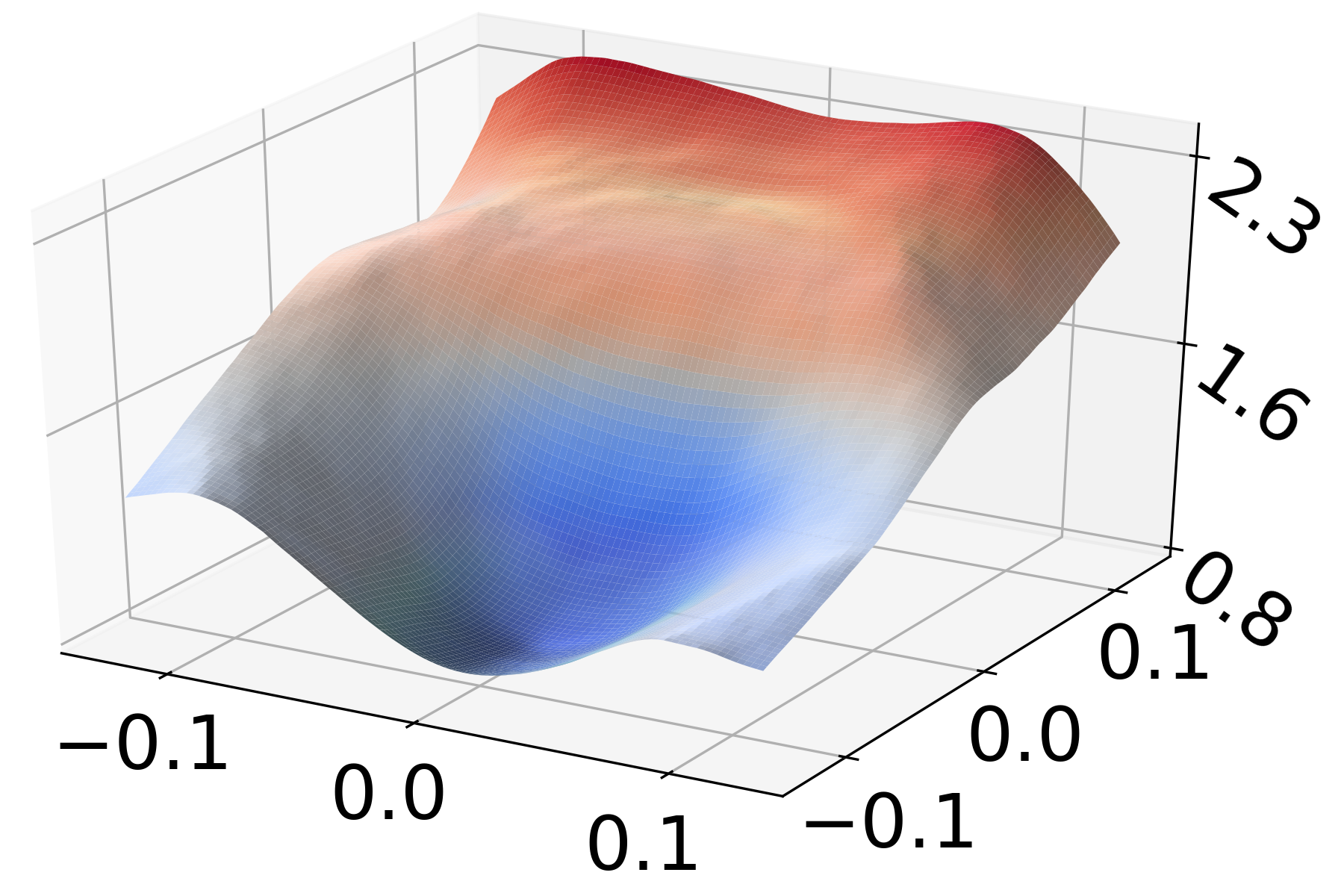}\\
\end{tabularx}
    \caption{Comparison of loss landscapes of PreRN-18 models trained by AT (the first row) and our methods (the
second and third row). Loss plots in each column are generated from the same original image randomly chosen from the
CIFAR-10 test set. z-axis denotes the loss value. Following the setting in~\cite{engstrom2018evaluating}, we plot the loss landscape function: $z=loss(x\cdot r_{1}+y\cdot  r_{2})$ where $r_{1}=sign(\nabla_{x}f(x))$ and $r_2 \sim Rademacher(0.5)$. }
\label{Fig:losslandscape_input}
\vskip -0.1 in
\end{figure*}

\subsection{Ablations and Visualizations}
\label{sec:ablation}
In this section, we first conduct ablation studies to show the effectiveness of the designed optimization trajectories and the unseen hold-out set in WOT. Then we investigate the impact of gaps:$m$ and the number of gaps:$k$, the effect of WOT on the loss landscapes w.r.t weight space, and the visualization of $\alpha$ for blocks. The results are shown in Table~\ref{tab:extrabaslines_ciar10}, Figure~\ref{fig:impactofhyper}, and Figure~\ref{fig:landscapes_weights}. All experiments in the two figures are conducted on CIFAR-10 with PreRN-18 based on AT except for Figure~\ref{fig:landscapes_weights} where Trades is also included. The robust accuracy is evaluated under AA-$L_{\infty}$ attack for all three figures.

\textbf{Ablation studies} To demonstrate the effectiveness of the designed optimization trajectories and the unseen hold-out set in boosting adversarial robustness, we designed the following baselines: 1) Keep the same unseen hold-out set and training strategy with WOT but optimize model weights instead of $\alpha$ on the unseen hold-out set (Abbreviated as ``B1" ); 2) Keep the same unseen hold-out set and optimize the hyperparameter of SWA by the hold-out set (Abbreviated as ``B2" ); 3) Replace the unseen hold-out set with a seen set, i.e. keep the same number of samples from the training set (Abbreviated as ``B3" ).  Results in Table~\ref{tab:extrabaslines_ciar10} show that \textbf{\circled{1}} AT+WOT-W/B outperforms AT+B1 and AT+B2, indicating the designed optimization trajectories play key roles in WOT. \textbf{\circled{2}} AT+WOT-W/B outperforms AT+B3 with a large margin, indicating the unseen hold-out set is crucial for WOT.

\textbf{Impact of $m$ and $k$.} Figure~\ref{fig:impactofhyper} shows the impact of gaps $m$ and number of gaps $k$ on robust accuracy under AA-$L_{\infty}$ attack. In the left figure, we observe that robust accuracy increases with an increase of $m$. Besides, we find that WOT-W is more sensitive to $m$ than WOT-B. The right figure shows that both WOT-W and WOT-B are not sensitive to the number of gaps $k$.

\textbf{Averaged $\alpha$ for Blocks.} To shed insights on why WOT-B outperforms WOT-W, we plot the learned $\alpha$ for each block. Experiments are conducted on CIFAR-10 with PreRN-18 based on WOT-B (K=4, m=400). Results in Figure~\ref{fig:landscapes_weights:c} and Figure~\ref{fig:landscapes_weights:d} show that the magnitude of learned $\alpha$ are  different among blocks. Specifically, WOT-B assigns a large value of $\alpha$  for middle blocks, i.e., Block-2,3,4,5, and a small value of $\alpha$ for the bottom and top blocks, i.e., Block-1,6. This indicates that assigning different weights for different blocks may play a crucial role in boosting adversarial robustness.

\textbf{Visualizing loss landscape.} We expect WOT  to search flatter minima for adversarial training to boost its robust generalization. We demonstrate that it indeed happens via visualizing the loss landscape with respect to  weight space (Figure~\ref{fig:landscapes_weights:a} and Figure~\ref{fig:landscapes_weights:b}) and input space (Figure~\ref{Fig:losslandscape_input}). Figure~\ref{fig:landscapes_weights:a} and Figure~\ref{fig:landscapes_weights:b} show that WOT+baseline obtains flatter minima than baseline, which indicates an improved robust generalization~\cite{stutz2021relating,wu2020adversarial}. Figure~\ref{Fig:losslandscape_input} demonstrates that, compared to AT, WOT achieves a lower loss value as the x-axis and y-axis values increase, indicating a lower curvature. This finding is consistent with the robust generalization claim presented in~\cite{Moosavi-Dezfooli2018}.



\section{Conclusion}
In this paper, we propose a new method named weighted optimization trajectories (WOT) for improving adversarial robustness and avoiding robust overfitting. We re-weight the optimization trajectories in time by maximizing the robust performance on an unseen hold-out set during the training process. The comprehensive experiments demonstrate: (1) WOT can effectively improve adversarial robustness across various adversarial training variants, model architectures, and benchmark datasets. (2) WOT enjoys superior performance in mitigating robust overfitting.  Moreover, visualizing analysis validates that WOT flattens the loss landscape with respect to input and weight space, showing an improved robust generalization. 

\section{Acknowledgement}
This work is partially supported by NWO EDIC and EDF KOIOS projects.  Part of this work used the Dutch national e-infrastructure with the support of the SURF Cooperative using grant no. NWO2021.060, EINF-5587 and EINF-5587/L1. We would like to express our deepest gratitude to the anonymous reviewers whose insightful comments and suggestions significantly improved the quality of this paper.

\section{Ethical Statement}
Our proposed method, Weighted Optimization Trajectories (WOT), enhances the robustness of deep neural networks against adversarial attacks. While the primary goal is to improve AI system security, it is crucial to consider the ethical implications of our work:
\begin{itemize}
    \item Personal Data Protection: Researchers and practitioners must ensure proper data handling, privacy, and compliance with data protection laws when working with personal data.
    \item Privacy Preservation: Improved robustness could inadvertently increase the capacity to infer personal information from data. Privacy-preserving techniques, like differential privacy, should be employed to mitigate these risks.
\end{itemize}

%
%
%
\bibliographystyle{splncs04}
\bibliography{splncs04}

\begin{thebibliography}{10}
\providecommand{\url}[1]{\texttt{#1}}
\providecommand{\urlprefix}{URL }
\providecommand{\doi}[1]{https://doi.org/#1}

\bibitem{alayrac2019labels}
Alayrac, J.B., Uesato, J., Huang, P.S., Fawzi, A., Stanforth, R., Kohli, P.:
  Are labels required for improving adversarial robustness? Advances in Neural
  Information Processing Systems  \textbf{32} (2019)

\bibitem{andriushchenko2020square}
Andriushchenko, M., Croce, F., Flammarion, N., Hein, M.: Square attack: a
  query-efficient black-box adversarial attack via random search. In: European
  Conference on Computer Vision. pp. 484--501. Springer (2020)

\bibitem{athalye2018obfuscated}
Athalye, A., Carlini, N., Wagner, D.: Obfuscated gradients give a false sense
  of security: Circumventing defenses to adversarial examples. arXiv preprint
  arXiv:1802.00420  (2018)

\bibitem{balaji2019instance}
Balaji, Y., Goldstein, T., Hoffman, J.: Instance adaptive adversarial training:
  Improved accuracy tradeoffs in neural nets. arXiv preprint arXiv:1910.08051
  (2019)

\bibitem{carlini2017towards}
Carlini, N., Wagner, D.: Towards evaluating the robustness of neural networks.
  In: 2017 IEEE Symposium on Security and Privacy (SP). pp. 39--57. IEEE (2017)

\bibitem{carmon2019unlabeled}
Carmon, Y., Raghunathan, A., Schmidt, L., Duchi, J.C., Liang, P.S.: Unlabeled
  data improves adversarial robustness. Advances in Neural Information
  Processing Systems  \textbf{32} (2019)

\bibitem{chen2020robust}
Chen, T., Zhang, Z., Liu, S., Chang, S., Wang, Z.: Robust overfitting may be
  mitigated by properly learned smoothening. In: International Conference on
  Learning Representations (2020)

\bibitem{chen2022sparsity}
Chen, T., Zhang, Z., Wang, P., Balachandra, S., Ma, H., Wang, Z., Wang, Z.:
  Sparsity winning twice: Better robust generaliztion from more efficient
  training. arXiv preprint arXiv:2202.09844  (2022)

\bibitem{cohen2019certified}
Cohen, J.M., Rosenfeld, E., Kolter, J.Z.: Certified adversarial robustness via
  randomized smoothing. arXiv preprint arXiv:1902.02918  (2019)

\bibitem{croce2020minimally}
Croce, F., Hein, M.: Minimally distorted adversarial examples with a fast
  adaptive boundary attack. In: International Conference on Machine Learning.
  pp. 2196--2205. PMLR (2020)

\bibitem{croce2020reliable}
Croce, F., Hein, M.: Reliable evaluation of adversarial robustness with an
  ensemble of diverse parameter-free attacks. In: International conference on
  machine learning. pp. 2206--2216. PMLR (2020)

\bibitem{deng2009imagenet}
Deng, J., Dong, W., Socher, R., Li, L.J., Li, K., Fei-Fei, L.: Imagenet: A
  large-scale hierarchical image database. In: 2009 IEEE conference on computer
  vision and pattern recognition. pp. 248--255. Ieee (2009)

\bibitem{dong2019evading}
Dong, Y., Pang, T., Su, H., Zhu, J.: Evading defenses to transferable
  adversarial examples by translation-invariant attacks. In: Proceedings of the
  IEEE/CVF Conference on Computer Vision and Pattern Recognition. pp.
  4312--4321 (2019)

\bibitem{dong2021exploring}
Dong, Y., Xu, K., Yang, X., Pang, T., Deng, Z., Su, H., Zhu, J.: Exploring
  memorization in adversarial training. arXiv preprint arXiv:2106.01606  (2021)

\bibitem{dziugaite2017computing}
Dziugaite, G.K., Roy, D.M.: Computing nonvacuous generalization bounds for deep
  (stochastic) neural networks with many more parameters than training data.
  arXiv preprint arXiv:1703.11008  (2017)

\bibitem{elisseeff2005stability}
Elisseeff, A., Evgeniou, T., Pontil, M., Kaelbing, L.P.: Stability of
  randomized learning algorithms. Journal of Machine Learning Research
  \textbf{6}(1) (2005)

\bibitem{engstrom2018evaluating}
Engstrom, L., Ilyas, A., Athalye, A.: Evaluating and understanding the
  robustness of adversarial logit pairing. arXiv preprint arXiv:1807.10272
  (2018)

\bibitem{foret2020sharpness}
Foret, P., Kleiner, A., Mobahi, H., Neyshabur, B.: Sharpness-aware minimization
  for efficiently improving generalization. arXiv preprint arXiv:2010.01412
  (2020)

\bibitem{Girshick2014}
Girshick, R., Donahue, J., Darrell, T., Malik, J.: {Rich feature hierarchies
  for accurate object detection and semantic segmentation}. In: Proceedings of
  the IEEE Computer Society Conference on Computer Vision and Pattern
  Recognition (2014). \doi{10.1109/CVPR.2014.81}

\bibitem{Goodfellow2014}
Goodfellow, I.J., Shlens, J., Szegedy, C.: {Explaining and Harnessing
  Adversarial Examples}  (dec 2014), \url{http://arxiv.org/abs/1412.6572}

\bibitem{hardt2016train}
Hardt, M., Recht, B., Singer, Y.: Train faster, generalize better: Stability of
  stochastic gradient descent. In: International conference on machine
  learning. pp. 1225--1234. PMLR (2016)

\bibitem{He2016}
He, K., Zhang, X., Ren, S., Sun, J.: {Deep residual learning for image
  recognition}. In: Proceedings of the IEEE conference on computer vision and
  pattern recognition. pp. 770--778 (2016)

\bibitem{Hinton2012}
Hinton, G., Deng, L., Yu, D., Dahl, G.E., Mohamed, A.r., Jaitly, N., Senior,
  A., Vanhoucke, V., Nguyen, P., Sainath, T.N., Kingsbury, B.: {Deep Neural
  Networks for Acoustic Modeling in Speech Recognition}. Ieee Signal Processing
  Magazine  (2012). \doi{10.1109/MSP.2012.2205597}

\bibitem{huang2020bridging}
Huang, T., Menkovski, V., Pei, Y., Pechenizkiy, M.: Bridging the performance
  gap between fgsm and pgd adversarial training. arXiv preprint
  arXiv:2011.05157  (2020)

\bibitem{huang2021calibrated}
Huang, T., Menkovski, V., Pei, Y., Pechenizkiy, M.: calibrated adversarial
  training. In: Asian Conference on Machine Learning. pp. 626--641. PMLR (2021)

\bibitem{huang2022direction}
Huang, T., Menkovski, V., Pei, Y., Wang, Y., Pechenizkiy, M.:
  Direction-aggregated attack for transferable adversarial examples. ACM
  Journal on Emerging Technologies in Computing Systems (JETC)  \textbf{18}(3),
   1--22 (2022)

\bibitem{jiang2019fantastic}
Jiang, Y., Neyshabur, B., Mobahi, H., Krishnan, D., Bengio, S.: Fantastic
  generalization measures and where to find them. arXiv preprint
  arXiv:1912.02178  (2019)

\bibitem{krizhevsky2010cifar}
Krizhevsky, A., Nair, V., Hinton, G.: Cifar-10 (canadian institute for advanced
  research). URL http://www. cs. toronto. edu/kriz/cifar. html  \textbf{5}
  (2010)

\bibitem{madry2017towards}
Madry, A., Makelov, A., Schmidt, L., Tsipras, D., Vladu, A.: Towards deep
  learning models resistant to adversarial attacks. arXiv preprint
  arXiv:1706.06083  (2017)

\bibitem{Madry2017}
Madry, A., Makelov, A., Schmidt, L., Tsipras, D., Vladu, A.: {Towards Deep
  Learning Models Resistant to Adversarial Attacks}  (jun 2017),
  \url{http://arxiv.org/abs/1706.06083}

\bibitem{Moosavi-Dezfooli2016}
Moosavi-Dezfooli, S.M., Fawzi, A., Frossard, P.: {DeepFool: A Simple and
  Accurate Method to Fool Deep Neural Networks}. In: Proceedings of the IEEE
  Computer Society Conference on Computer Vision and Pattern Recognition. vol.
  2016-Decem, pp. 2574--2582. IEEE Computer Society (dec 2016).
  \doi{10.1109/CVPR.2016.282}

\bibitem{Moosavi-Dezfooli2018}
Moosavi-Dezfooli, S.M., Fawzi, A., Uesato, J., Frossard, P.: {Robustness via
  curvature regularization, and vice versa}  (nov 2018),
  \url{http://arxiv.org/abs/1811.09716}

\bibitem{netzer2011reading}
Netzer, Y., Wang, T., Coates, A., Bissacco, A., Wu, B., Ng, A.Y.: Reading
  digits in natural images with unsupervised feature learning  (2011)

\bibitem{pang2022robustness}
Pang, T., Lin, M., Yang, X., Zhu, J., Yan, S.: Robustness and accuracy could be
  reconcilable by (proper) definition. arXiv preprint arXiv:2202.10103  (2022)

\bibitem{Qin2019}
Qin, C., Martens, J., Gowal, S., Krishnan, D., Krishnamurthy, Dvijotham, Fawzi,
  A., De, S., Stanforth, R., Kohli, P.: {Adversarial Robustness through Local
  Linearization}  (jul 2019), \url{http://arxiv.org/abs/1907.02610}

\bibitem{rebuffi2021fixing}
Rebuffi, S.A., Gowal, S., Calian, D.A., Stimberg, F., Wiles, O., Mann, T.:
  Fixing data augmentation to improve adversarial robustness. arXiv preprint
  arXiv:2103.01946  (2021)

\bibitem{rhu2016vdnn}
Rhu, M., Gimelshein, N., Clemons, J., Zulfiqar, A., Keckler, S.W.: vdnn:
  Virtualized deep neural networks for scalable, memory-efficient neural
  network design. In: 2016 49th Annual IEEE/ACM International Symposium on
  Microarchitecture (MICRO). pp. 1--13. IEEE (2016)

\bibitem{rice2020overfitting}
Rice, L., Wong, E., Kolter, Z.: Overfitting in adversarially robust deep
  learning. In: International Conference on Machine Learning. pp. 8093--8104.
  PMLR (2020)

\bibitem{ross2018improving}
Ross, A., Doshi-Velez, F.: Improving the adversarial robustness and
  interpretability of deep neural networks by regularizing their input
  gradients. In: Proceedings of the AAAI Conference on Artificial Intelligence.
  vol.~32 (2018)

\bibitem{schmidt2018adversarially}
Schmidt, L., Santurkar, S., Tsipras, D., Talwar, K., Madry, A.: Adversarially
  robust generalization requires more data. In: Advances in Neural Information
  Processing Systems. pp. 5014--5026 (2018)

\bibitem{sehwag2021robust}
Sehwag, V., Mahloujifar, S., Handina, T., Dai, S., Xiang, C., Chiang, M.,
  Mittal, P.: Robust learning meets generative models: Can proxy distributions
  improve adversarial robustness? arXiv preprint arXiv:2104.09425  (2021)

\bibitem{simonyan2014very}
Simonyan, K., Zisserman, A.: Very deep convolutional networks for large-scale
  image recognition. arXiv preprint arXiv:1409.1556  (2014)

\bibitem{singla2021low}
Singla, V., Singla, S., Feizi, S., Jacobs, D.: Low curvature activations reduce
  overfitting in adversarial training. In: Proceedings of the IEEE/CVF
  International Conference on Computer Vision. pp. 16423--16433 (2021)

\bibitem{stutz2021relating}
Stutz, D., Hein, M., Schiele, B.: Relating adversarially robust generalization
  to flat minima. In: Proceedings of the IEEE/CVF International Conference on
  Computer Vision. pp. 7807--7817 (2021)

\bibitem{Szegedy2013}
Szegedy, C., Zaremba, W., Sutskever, I., Bruna, J., Erhan, D., Goodfellow, I.,
  Fergus, R.: {Intriguing properties of neural networks}  (dec 2013),
  \url{http://arxiv.org/abs/1312.6199}

\bibitem{uesato2018adversarial}
Uesato, J., O’donoghue, B., Kohli, P., Oord, A.: Adversarial risk and the
  dangers of evaluating against weak attacks. In: International Conference on
  Machine Learning. pp. 5025--5034. PMLR (2018)

\bibitem{wang2019improving}
Wang, Y., Zou, D., Yi, J., Bailey, J., Ma, X., Gu, Q.: Improving adversarial
  robustness requires revisiting misclassified examples. In: International
  Conference on Learning Representations (2020)

\bibitem{wu2020revisiting}
Wu, D., Wang, Y., Xia, S.t.: Revisiting loss landscape for adversarial
  robustness. arXiv preprint arXiv:2004.05884  (2020)

\bibitem{wu2020adversarial}
Wu, D., Xia, S.T., Wang, Y.: Adversarial weight perturbation helps robust
  generalization. Advances in Neural Information Processing Systems
  \textbf{33} (2020)

\bibitem{xie2019improving}
Xie, C., Zhang, Z., Zhou, Y., Bai, S., Wang, J., Ren, Z., Yuille, A.L.:
  Improving transferability of adversarial examples with input diversity. In:
  Proceedings of the IEEE/CVF Conference on Computer Vision and Pattern
  Recognition. pp. 2730--2739 (2019)

\bibitem{yu2021robust}
Yu, C., Han, B., Gong, M., Shen, L., Ge, S., Du, B., Liu, T.: Robust weight
  perturbation for adversarial training  (2021)

\bibitem{zagoruyko2016wide}
Zagoruyko, S., Komodakis, N.: Wide residual networks. arXiv preprint
  arXiv:1605.07146  (2016)

\bibitem{zhang2017musings}
Zhang, C., Liao, Q., Rakhlin, A., Sridharan, K., Miranda, B., Golowich, N.,
  Poggio, T.: Musings on deep learning: Properties of sgd. Tech. rep., Center
  for Brains, Minds and Machines (CBMM) (2017)

\bibitem{zhang2019theoretically}
Zhang, H., Yu, Y., Jiao, J., Xing, E., El~Ghaoui, L., Jordan, M.: Theoretically
  principled trade-off between robustness and accuracy. In: International
  conference on machine learning. pp. 7472--7482. PMLR (2019)

\bibitem{zhang2020attacks}
Zhang, J., Xu, X., Han, B., Niu, G., Cui, L., Sugiyama, M., Kankanhalli, M.:
  Attacks which do not kill training make adversarial learning stronger. In:
  International conference on machine learning. pp. 11278--11287. PMLR (2020)

\bibitem{zhang2020geometry}
Zhang, J., Zhu, J., Niu, G., Han, B., Sugiyama, M., Kankanhalli, M.:
  Geometry-aware instance-reweighted adversarial training. arXiv preprint
  arXiv:2010.01736  (2020)

\bibitem{zhou2018generalization}
Zhou, Y., Liang, Y., Zhang, H.: Generalization error bounds with probabilistic
  guarantee for sgd in nonconvex optimization. arXiv preprint arXiv:1802.06903
  (2018)

\end{thebibliography}
%





\clearpage
\appendix
\section{Implementations details}\label{apped_imple}
The implementation of WOT is based on the PyTorch library and all experiments in this study are run on a single A100 GPU.  Following~\cite{chen2020robust}, SWA starts after the first learning decay. All models used in this study are trained with $\epsilon=8/255.$

All baselines are achieved by running the code provided by their authors (Table~\ref{tab:download_links}).

\begin{table*}[!h]
\caption{Download Links for baseline codes.}
\begin{center}
\begin{small}
\begin{sc}
\begin{adjustbox}{width=1.0\textwidth}
\begin{tabular}{l|c}
\toprule
Methods & download links \\
\midrule
AT+early stop &\url{https://github.com/locuslab/robust_overfitting.git} \\
Trades &\url{https://github.com/yaodongyu/TRADES.git}\\
MART &\url{https://github.com/YisenWang/MART.git}\\
AWP &\url{https://github.com/csdongxian/AWP.git}\\
AT+SWA&\url{https://github.com/VITA-Group/Alleviate-Robust-Overfitting.git}\\
FAT &\url{https://github.com/zjfheart/Friendly-Adversarial-Training.git}\\
\bottomrule
\end{tabular}
\end{adjustbox}
\end{sc}
\end{small}
\end{center}
\label{tab:download_links}
\end{table*}

\section{Block Details for WOT-B}\label{apped_blocks}
We naturally divide ResNet kinds of  architectures according to their original block design. For VGG-16,  we group the layers with the same number of channels as a block. We share the details in Table~\ref{tab:blocks}.
 
\begin{table*}[htb]
\caption{The corresponding relationship between layers and blocks for PreActResNet-18, WRN-34-10 and VGG-16.}
\begin{center}
\begin{small}
\begin{sc}
\begin{adjustbox}{width=1.0\textwidth}
\begin{tabular}{cc|cc|cc}
\toprule
\multicolumn{2}{c|}{PreActResNet-18} &\multicolumn{2}{|c|}{WRN-34-10}&\multicolumn{2}{|c}{VGG-16}\\
\midrule
Layers & Block Index &Layers & Block Index&Layers & Block Index \\
1 & 1&1 &1&1-2&1\\
2-5&2&2-12&2&3-4&2\\
6-9&3&13-23&3&5-7&3\\
10-13&4&24-34&4&8-13&4\\
14-17&5 &Final linear Layer&5&13-16&5\\
Final linear Layer&6&-&-&-&-\\
\bottomrule
\end{tabular}
\end{adjustbox}
\end{sc}
\end{small}
\end{center}
\label{tab:blocks}
\end{table*}

\section{Pseudocode} \label{apped_pso}
The Pseudocode can be found in Algorithm~\ref{alg:MO}.
\begin{algorithm}[H]
\small
  \caption{WOT for AT-PGD}
  \label{alg:MO}
\begin{algorithmic}[1]
  \STATE {\bfseries Input:} A model $f_{w}(x)$ with parameters $w$, training set $(x,y)\in S$, unseen set $(x_{uns},y_{uns}) \in S_{uns}$, total training steps T, WOT starting epoch:$p$, Gaps: $m$, Number of Gaps:$k$, and adversarial training loss $L(\cdot)$. 
  \STATE {\bfseries Output:} $f_{w}(x)$ with trained parameters. 
  \STATE initialize model weights $w$, $\Delta W=\emptyset$. 
  \FOR{$t=1$ {\bfseries to} $T$}
      \STATE \# Normal adversarial training for $f_{w}$.
      \STATE $w_{(t)}=w_{(t-1)}-lr\cdot \nabla_{w}{L(f_{w_{(t-1)}}(x+\Delta x),y)}$
      \STATE \# WOT starts after p epochs (steps).
      \IF{$t>p$} 
      \IF{($t-p$)\%$m$=0}
      \STATE \#Collect optimization trajectories.
      \STATE $\Delta w =w_{(t)}-w_{(t-m)}$
      \STATE $\Delta W =\Delta W \cup \{\Delta w\}$
      \ENDIF
      \IF{($t-p$)\%($m*k$)=0}
      \STATE \#Optimize $\alpha$ based on Eq.~\ref{update_alpha_m} and Eq.~\ref{update_alpha}
      \STATE Initialize $\alpha$ with zero.
      \FOR{$i=1$ {\bfseries to} $n$}
            \STATE $m^{i}=m^{i-1}\cdot \gamma+\nabla_{\alpha}L(f_{\widetilde{w^{\prime}}}(x_{uns}+\Delta x_{uns}),y_{uns})$
            \STATE $\alpha=\alpha-lr\cdot m^{i}$
      \ENDFOR
      \STATE $w_{(t)}=w_{(t-mk)}+\widetilde{\Delta w}$
      \STATE $\Delta W=\emptyset$
      \ENDIF
      \ENDIF
  \ENDFOR
  \STATE Return $f_{w}(x)$
\end{algorithmic}
\end{algorithm}

\section{More Experiments}\label{appe_extraExp}
The experiments are conducted on CIFAR-10 based on FAT,  FAT+WOT-W, and FAT+WOT-B with PreActResNet-18 architecture. The results in Table~\ref{tab:more_experiments}  consistently verify that WOT can effectively improve adversarial robustness across multiple attacks without sacrificing of clean accuracy. Besides, we also report the robust performance of AT+EMA and compare it with AT+WOT-B/W. The results in Table~\ref{tab:extrabaslines_ema} show that AT+WOT-B/W outperforms AT+EMA  as well.

\begin{table*}[!h]
\caption{Test robustness under multiple adversarial attacks based on FAT. The experiments are conducted on CIFAR-10 with PreActResNet-18 architecture. The bold denotes the best performance.}
\begin{center}
\begin{small}
\begin{sc}
\begin{tabular}{l|ccccccc}
\toprule
Models & Clean &FGSM&PGD-20 & PGD-100&CW$_{\infty}$&AA-$L_{\infty}$\\
\midrule
FAT &86.90&54.186&44.69 &42.35 &44.84&40.71 \\
FAT+WOT-W(\textbf{Ours}) &87.25 &55.97 &47.81&45.71 &46.87&43.14\\
FAT+WOT-B(\textbf{Ours}) &\textbf{87.59}&\textbf{57.22} &\textbf{48.27} &\textbf{46.03}&\textbf{47.62}&\textbf{43.40}\\
\bottomrule
\end{tabular}
\end{sc}
\end{small}
\end{center}
\label{tab:more_experiments}
\end{table*}

\begin{table}[tbh]
\caption{The robust accuracy of EMA VS WOT on CIFAR-10 with PreActResNet-18 architecture}
        \begin{center}
\begin{small}
\begin{sc}
\begin{adjustbox}{width=0.8\linewidth}
    \begin{tabular}{c|c|c|c|c}
    \toprule
          &PGD-20 &PGD-100&CW$_\infty$&AA-L$_\infty$\\
         \midrule
         AT+EMA&52.88&52.24&50.75&47.94\\
         AT+WOT-B (m=400,k=4)&54.85&53.77&52.56&48.96\\
         AT+WOT-W (m=400,k=4)&53.19& 51.90 &51.74 &48.36\\
         \bottomrule
    \end{tabular}
    \end{adjustbox}
        \end{sc}
\end{small}
\end{center}
    \label{tab:extrabaslines_ema}
\end{table}

\section{Memory cost and Time cost}\label{limitations}
WOT takes more GPU memory space since it needs to record the optimization trajectories. We report the memory cost of WOT (Gaps:m=400, Number of Gaps: K=4) based on VGG-16 and PreActResNet-18 respectively in Table~\ref{tab:momery}. It shows that even though we need extra memory to cache $\Delta w$, the memory overhead in practice is quite small since the majority of memory usage is concentrated on the intermediate feature maps and gradient maps, accounting for 81\% of memory usage on AlexNet and 96\% on VGG-16 for an example~\cite{rhu2016vdnn}.  

Besides, WOT consumes more computation time since it needs to optimize the $\alpha$ by maximizing the robust performance on an unseen set. However, the extra computation time cost is negligible since the unseen set is very small. For example, AT+WOT takes \textbf{135} minutes to train PreActResNet-18 model for 100 epochs while  AT takes \textbf{129} minutes to train the model for 100 epochs. 

\begin{table}[H]
\caption{Momory cost of WOT-B (Gaps:400, Number of Gaps:K=4)}
    \begin{center}
\begin{small}
\begin{sc}
    \begin{tabular}{c|c|c}
    \toprule
         &WOT-B &Vanilla AT\\
         \midrule
          VGG&3771M&3409M \\ 
          PreAcResNet-18&4635M&4213M\\
         \bottomrule
    \end{tabular}
    \end{sc}
\end{small}
\end{center}
    \label{tab:momery}
\end{table}

\section{The rationality for constraining $\alpha$ to $[0,1]$} \label{rationalityofalpha}
 We constrain $\alpha$ to [0, 1] such that the refined optimization trajectories will not go too far away from the optimization trajectories.  In contrast, if we do not constrain $\alpha$, refining optimization trajectories could lead to worse performance or even cause collapse.

We conduct experiments on CIFAR-10 and CIFAR-100 with PreActResNet-18 model to show that constraining $\alpha$ to $[0,1]$ is a reasonable choice. The experimental settings are as follows:
\begin{itemize}
    \item WOT-B without constraining $\alpha$. (Abbreviated as WOT-B (No constraints))
    \item WOT-B with sum($\alpha$) =1.
    \item WOT-B with max($\alpha$)=0.5, 0.8, and $\alpha \in [0,$max$(\alpha)]$.
\end{itemize}
The results are reported in Table~\ref{tab:alpha_setting} where we can see that without our constraint, there is worse performance. We empirically find that neither constraining sum($\alpha$) =1 nor constraining max($\alpha$) with a smaller value ($<1$)  can match the performance of our default setting. 

\begin{table}[H]
\caption{Robust Accuracy under PGD-10 attack for WOT, WOT without constraints for $\alpha$, WOT with the constraint by setting sum($\alpha$) =1 or max($\alpha$)=0.5, 0.8. Number of Gaps:k=4. NaN denotes the refining process leads to a NaN loss.}
        \begin{center}
\begin{sc}
\begin{adjustbox}{width=0.8\textwidth}
    \begin{tabular}{c|c|c}
    \toprule
         Methods &CIFAR-10 &CIFAR-100\\
            \midrule
          WOT-B (Gaps:m=400)&55.83& 30.22\\
          WOT-B (Gaps:m=800)&\textbf{56.22}& \textbf{30.47}\\
         WOT-B (Sum($\alpha$)=1, Gaps:m=400)&53.12&28.43\\
         WOT-B (max($\alpha$)=0.8, Gaps:m=400)&55.18&29.83\\
         WOT-B (max($\alpha$)=0.5, Gaps:m=400)&55.34&29.66\\
         WOT-B (NO constraints, Gaps:m=400)&55.68&NaN\\
         \bottomrule
    \end{tabular}
    \end{adjustbox}
        \end{sc}
\end{center}
    \label{tab:alpha_setting}
\end{table} 
\section{The effect of the size of the hold-out set }
We conduct experiments to show the effect of the size of the hold-out set (validation set)  based on CIFAR-10 with PreActResNet-18 and WOT-B (Gaps:400, Number of Gaps:k=4).  The results are reported below in Table~\ref{tab:size} and show that the robust accuracy decreases with the size increasing from 1000 to 8000. We conjecture that a large size of the hold-out set  reduces the size of the training set, which would deteriorate the optimizing trajectories, leading to worse performance.  

\begin{table}[H]
\caption{Robust and clean accuracy under PGD-10 on CIFAR-10 with various sizes of the unseen hold-out set.}
        \begin{center}
\begin{small}
\begin{sc}
    \begin{tabular}{c|c|c}
    \toprule
         size&Clean Accuracy &Robust Accuracy\\
         \midrule
         1000&83.97 &\textbf{55.83} \\ 
         2000&\textbf{84.30}&55.56\\
         4000&83.90&55.10\\
         8000&84.17 &54.58\\
         \bottomrule
    \end{tabular}
            \end{sc}
\end{small}
\end{center}
    \label{tab:size}
\end{table} 

\section{The influence of the choice of the validation set}
We report the standard deviation of robust accuracy among the three repeated runs based on different validation sets on CIFAR-10 and CIFAR-100 respectively with WOT-B. The robust accuracy is calculated under CW$_\infty$ attack. The different validation sets are randomly sampled from CIFAR-10/100 with a different seed. 

The results (In Table~\ref{tab:diffval}) show that the standard deviations are less than 0.3\%, in line with the statement in our paper.

\begin{table}[H]
\caption{The standard deviations on the robust accuracy with three runs using different samples in the validation set.}
        \begin{center}
\begin{small}
\begin{sc}
    \begin{tabular}{c|c|c}
    \toprule
         Validation set&CIFAR-10 & CIFAR-100\\
         \midrule
         set 1&52.01 &27.04 \\ 
         set 2&52.36&27.14\\
         set 3&52.11&27.3\\\
         mean &52.16 &27.16\\
         std &0.18 & 0.13\\
         \bottomrule
    \end{tabular}
    \end{sc}
\end{small}
\end{center}
    \label{tab:diffval}
\end{table}

\section{Connection to SWA}\label{ConSWA}
Given a deep model $f_{w}$ with parameters $w$. The parameter of $f$ are denoted as $w_{0}$, $w_{1}$, $w_{2}$,...,$w_{n}$ at $T_{0}$, $T_{1}$, $T_{2}$,..,$T_{n}$ epoches respectively. The gradients of the training loss w.r.t $w$ are denoted as $g_{0}$, $g_{1}$, $g_{2}$,...,$g_{n}$ at $T_{0}$, $T_{1}$, $T_{2}$,..,$T_{n}$ epoches. The learning rate and momentum decay factor are denoted as $\lambda$ and $\alpha$ respectively. The momentum buffer at epoch $T_0$ is given $M_0$.

\textbf{Momentum can be expressed as follows:}
\begin{align}
    M_{t}=M_{t-1}\cdot \alpha+g_{t-1}
\end{align}

\textbf{Then, parameters updating can be formulated as follows:}
\begin{align}
    w_{1}&=w_{0}-\lambda \cdot M_1 \\
    w_{2}&=w_{0}-\lambda \cdot (M_1+M_2)\\
    w_{t}&=w_{0}-\lambda \cdot (M_1+...+M_t) 
\end{align}

\textbf{SWA: Averaging model weights along the trajectory of optimization. We expand it by gradients:} 
\begin{align}
    w_{swa}&=\frac{\sum_{t=0}^{t=n-1}{w_{t}}}{n}\\
    w_{swa}&=\frac{n\cdot w_{0}-\lambda \cdot \{(n-1)\cdot M_1+(n-2)\cdot M_2+...+M_{n-1}\}}{n} \\
    &=w_{0}-\lambda \cdot (\frac{n-1}{n}\cdot M_1+\frac{n-2}{n}\cdot M_2 +...\frac{1}{n} M_{n-1})
\end{align}

\begin{align}
w_{swa}-w_{0} &=-\lambda \cdot (\frac{n-1}{n}\cdot M_1+\frac{n-2}{n}\cdot M_2 +...\frac{1}{n} M_{n-1})\\ 
&= -\lambda \cdot (M_{0}(\frac{n-1}{n}\cdot\alpha+\frac{n-2}{n}\cdot\alpha^2+\frac{n-3}{n}\cdot\alpha^3 +...+\frac{1}{n} \cdot \alpha^{n-1})\\
&+g_{0}\cdot(\frac{n-1}{n}+\frac{n-2}{n}\cdot\alpha+\frac{n-3}{n}\cdot\alpha^2 +...+\frac{1}{n}\alpha^{n-2})\\ &+g_{1}\cdot(\frac{n-2}{n}+\frac{n-3}{n}\cdot\alpha+\frac{n-4}{n}\cdot\alpha^2 +...+\frac{1}{n}\alpha^{n-3})+...\\
&+\frac{1}{n}g_{n-2})
\end{align}

From the analysis above, SWA heuristically re-weights the gradients in essence. If we set gaps $m$ to 1 batch optimization step and a number of $gaps$ equals the length from the checkpoint that SWA starts to the checkpoint that SWA ends, WOT equals to re-weight the gradients with learned weights.

\end{document}